\documentclass[journal]{IEEEtran}

\usepackage{bbm}
\usepackage{epsfig}
\usepackage{amssymb} 
\usepackage{latexsym}
\usepackage{euscript}
\usepackage{amsmath}
\usepackage{amsthm}
\usepackage[numbers]{natbib}
\usepackage{notoccite}
\usepackage{float}
\usepackage[english]{babel}
\usepackage{fancyhdr}
\usepackage{lastpage}
\usepackage{amssymb}
\usepackage{mathtools}
\newtheorem{remark}{Remark}
\usepackage{scalerel}
\usepackage{array}
\newcommand{\params}[0]{\Psi}
 
\newcommand{\func}[2]{\ensuremath{{#1}^{#2}}}
\newcommand{\matidx}[2]{\ensuremath{[{#1}]_{#2}}}
\newcommand{\genscalar}[0]{\ensuremath{e}}
\newcommand{\genvec}[0]{\ensuremath{\mathbf{e}}}
\newcommand{\genmat}[0]{\ensuremath{E}}
\usepackage{graphicx}
\usepackage{easybmat}
\usepackage{multirow,bigdelim}
\usepackage{tikz}
\usetikzlibrary{matrix,decorations.pathreplacing}

\makeatletter
\DeclareFontFamily{U}{tipa}{}
\DeclareFontShape{U}{tipa}{m}{n}{<->tipa10}{}
\newcommand{\arc@char}{{\usefont{U}{tipa}{m}{n}\symbol{62}}}%

\newcommand{\arc}[1]{\mathpalette\arc@arc{#1}}

\newcommand{\arc@arc}[2]{%
  \sbox0{$\m@th#1#2$}%
  \vbox{
    \hbox{\resizebox{\wd0}{\height}{\arc@char}}
    \nointerlineskip
    \box0
  }%
}
\makeatother
\ifCLASSOPTIONcompsoc
    \usepackage[caption=false, font=normalsize, labelfont=sf, textfont=sf]{subfig}
\else
    \usepackage[caption=false, font=footnotesize]{subfig}
\fi

\usepackage{calc}

\usepackage{accents}
\newcommand{\dbtilde}[1]{\accentset{\approx}{#1}}

\newcommand{\twoparts}[0]{\ensuremath{\alpha}}

\begin{document}
\title{Node-Centric Graph Learning from Data\\for Brain State Identification}
\author{Nafiseh Ghoroghchian, David M. Groppe, Roman Genov,~\IEEEmembership{Senior~Member,~IEEE},\\ Taufik A. Valiante, and Stark C. Draper,~\IEEEmembership{Senior~Member,~IEEE}\\
\thanks{This work was supported in part by the Natural Sciences and Engineering Research Council (NSERC) of Canada, including through a Discovery Research Grant.}  
\thanks{N. Ghoroghchian, R. Genov and S.~C.~Draper are with the Edward S. Rogers Sr. Dept. of Electrical and Computer Engineering, University of Toronto, Toronto, ON M5S 3G4, Canada (e-mails: nafiseh.ghoroghchian@mail.utoronto.ca, roman@eecg.utoronto.ca, stark.draper@utoronto.ca).}
\thanks{D. M. Groppe is with the Krembil Research Institute, Toronto,
ON M5T 2S8, Canada. (email: david.groppe@uhnresearch.ca)}
\thanks{T. A. Valiante is with the  Institute of Biomaterials and Biomedical Engineering, University of Toronto; Dept. of Electrical and Computer Engineering, University of Toronto; Krembil Research Institute, Clinical and Computational Neuroscience, Toronto Western Hospital, ON M5T 2S8, Canada (email: Taufik.Valiante@uhn.ca).}
}

\markboth{Published in IEEE Transactions on Signal and Information Processing over Networks, 2020.}{}

\maketitle

\begin{abstract}
Data-driven graph learning models a network by determining the strength of connections between its nodes. The data refers to a graph signal which associates a value with each graph node. Existing  graph learning methods either use simplified models for the graph signal, or they are prohibitively expensive in terms of computational and memory requirements. This is particularly true when the number of nodes is high or there are temporal changes in the network. In order to consider  richer models with a reasonable computational tractability, we introduce a  graph learning method based on representation learning on graphs. Representation learning generates an embedding for each graph node, taking the information from neighbouring nodes into account. Our graph learning method further modifies the embeddings to compute the graph similarity matrix. In this work, graph learning is used to examine brain networks for brain state identification. We infer time-varying brain graphs from an extensive dataset of intracranial electroencephalographic (iEEG) signals from ten patients. We then apply the graphs as input to a classifier to distinguish seizure vs. non-seizure brain states. Using the binary classification metric of area under the receiver operating characteristic curve (AUC), this approach yields an average of 9.13 percent improvement when compared to two widely used brain network modeling methods.
\end{abstract}


%
\IEEEpeerreviewmaketitle

\section{Introduction}
\IEEEPARstart{I}{nferring} the relationship between nodes of a graph is an essential step to model structured data as a graph, and to facilitate data analysis and processing. Graphs represent a set of elements by nodes and the connections between elements by edges. The widespread use of graphs originates from their ability to represent interdependence in data by an underlying graphical model. Graph nodes are associated with data values, termed the \textit{signal on the graph}. For instance, if a bus transit network is considered as a graphical model, the set of passenger wait times at each bus stop is a graph signal. Another example is user information (e.g., age, interests) as the signal on the graph corresponding to a social network. Such graph-based signal modeling facilitates making inference about the signal. 

Graph learning  addresses the problem of building a graph by inferring the interrelationships in a signal \cite{dong2019learning}. The relationship between a pair of graph nodes typically describes their similarity. If an edge connects two nodes, the weight of the edge shows their similarity (connection strength). Sometimes, pairwise similarities in a graphical model are given explicitly, e.g., the strength of friendship between two persons in a friendship network. However, there are applications where the similarity must be derived from the signal. Examples of such data-driven graph learning include link prediction in social networks \cite{xiang2010modeling} and recommendation systems \cite{li2013recommendation}; protein-protein interaction networks \cite{airola2008graph}; network inference from signals in the human brain \cite{sporns2010networks}. 

There are numerous ways to infer a graph from a signal. Aside from conventional methods such as correlation, multiple methods have been introduced in the graph signal processing (GSP) literature using various models for graph signals \cite{egilmez2018graph,thanou2017learning,shen2017kernel}. A major downside to such  graph learning methods, which we will term edge-centric, for reasons we will later elaborate, is their lack of scalability. When the signal on a graph changes or the number of graph nodes increases, the new graph must be computed from scratch at high computational cost. As a result, scalability is an important factor in many applications, including time-varying brain networks.

\begin{figure}[!t]
\centering
\includegraphics[width=3.5in]{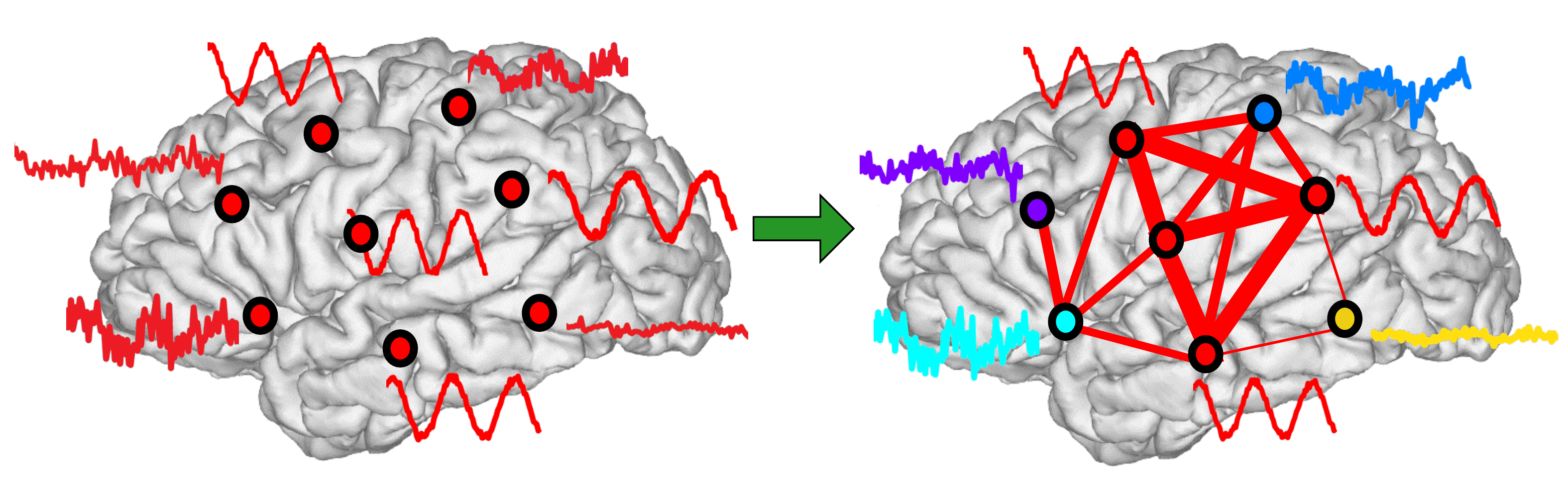}
\caption{An example of brain modeling as a graph. Initially, only a signal on the graph is available, without further information about the graph connections. The graph is then built using a data-driven graph learning method by quantifying statistical dependencies among the nodes. Thicker edges show stronger connections. }
\label{fig.brain_signal_mapping_graph}
\end{figure}

Our work is inspired by the network perspective of brain function. The brain consists of billions of neurons, connected to each other with biological wires called axons. Signal transmission along axons makes neuronal interactions possible and contributes to large-scale brain functions \cite{Valiante_2012}. Distinguishing different brain states, e.g., sleep vs. wake, seizure vs. non-seizure, remains a central problem to the basic understanding of the brain, development of disease biomarkers, and enhancing medical devices (e.g., anti-seizure neuromodulation \cite{o2018nurip}). Tools to describe the brain's activity as a network of interacting nodes are being developed \cite{sporns2010networks}. Such network descriptions of the brain are multivariate and may be better representations of brain states than simple univariate or bivariate approaches  \cite{sporns2010networks,ghoroghchian2018hierarchical}. 
Three widely known brain network models are structural, functional and effective connectivity  \cite{sporns2005human,friston2011functional,rubinov2010complex}. Structural connectivity refers to bundles of neurons, called white matter tracts, which has to do with physical links. Functional connectivity models undirected statistical relations among neuronal events. Effective connectivity is mostly about inferring directed causal neuronal relations \cite{friston2011functional}. At the macroscale, the functional and effective connectivities are often quantified using the signals acquired from electrodes (channels) placed in different brain regions. Fig.~\ref{fig.brain_signal_mapping_graph} depicts an example of such data-driven graph modeling. Functional connectivity is frequently computed using pairwise correlation \cite{khambhati2015dynamic,khambhati2017recurring} and pairwise coherence \cite{burns2014network,bastos2016tutorial}. Dynamic causal modeling is the most popular method to compute effective connectivity \cite{friston2011functional}.

The contributions of this paper are twofold.
First, motivated by drawbacks in the existing edge-centric graph learning methods, we propose a scalable node-centric graph learning scheme. We accomplish this by making use of representation learning on graphs. The method we developed is not only a rich model but has reasonable computational tractability. The tractability is achieved by using the notion of intra-graph generalization which enables reuse of the \textit{shared parameters} of a model to build new graphs. Second, we introduce a specific framework for brain signal processing. We apply the proposed  graph learning method to infer time-varying brain connectivity. 

The outline of the paper is as follows. In Sec.~\ref{sec:preliminaries}, the motivations behind proposing a node-centric perspective for graph learning are elaborated. Additionally, representation learning on graphs and one of its variants, GraphSAGE \cite{hamilton2017inductive}, are introduced. In order to address the two objectives of this work which differ from those of GraphSAGE, next in Sec.~\ref{sec:problem_statement}, we introduce the idea of a similarity matrix, make modifications to existing optimization problem, and derive a new problem formulation. We then apply the proposed  graph learning framework to brain network inference in Sec.~\ref{sec:brain_connectivity}. As an application of the proposed method for brain state identification, in Sec.~\ref{sec:results} we provide experimental results from an epilepsy dataset to classify pre-seizure and seizure vs. non-seizure brain states using the inferred brain graphs. Sec.~\ref{sec:conclusion} concludes the paper.

For reference, in Table~\ref{tab:notations} we list the main notation used throughout the paper. 

\begin{table}
\begin{center}
\caption{Table of Main Notations}
\label{tab:notations}
\begin{tabular}{|c|c|c|c|c|}
\hline \hline 
\textbf{Notation}  &  \textbf{Description} \\
\hline\hline
$A$ & adjacency matrix\\
\hline
$\mathcal{V}$ & set of all nodes\\
 \hline
$\mathcal{E}$ & set of all edges\\
\hline
$\mathcal{N}_u$ & neighbourhood of node $u$\\
\hline
$\mathbf{x}_u$ & graph signal of node $u$\\
\hline
$X$ & graph signal matrix\\
\hline
$T$ & length of the graph signal of each node\\
\hline
$S$ & similarity matrix\\
\hline
$\mathbf{z}_u$ & embedding of node $u$\\
\hline
$Z$ & nodes' embedding matrix\\
\hline
$\mathcal{I}$ & set of training samples' indices \\
\hline
$I$ & number of training samples \\
\hline
$\Psi$ & shared parameters in representation learning formulations\\
\hline
$\sigma$ & non-linear activation function\\
\hline
$\mathbf{h}^0_u$ & initial feature vector of node $u$ at iteration $k$\\
\hline
$\mathbf{h}^k_u$ & hidden feature vector of node $u$ at iteration $k$\\
\hline
 $g_u^k$ & aggregation function of node $u$ at iteration $k$\\
 \hline
$\Psi^k$ & shared parameters in  $g_u^k$s at iteration $k$\\
\hline
$D_0$&  length of a node's initial feature vector\\
\hline
$D$ & length of a node embedding, is equal to $2D_0$ \\
\hline
KL & Kullback-Leibler divergence\\
\hline
 $\func{f}{\textrm{NCDD}}$ & component-wise embedding-to-similarity mapping\\
\hline
$\mathbf{\Theta}$ &  set of parameters in the similarity definition\\
\hline
$q$ & node-wise mapping of signal to initial feature vector \\
\hline
 $\genvec$ &   universal notation for a vector \\
\hline
  $\genmat$ &   universal notation for a matrix \\
\hline
$\mathcal{W}$ & set of frequency bin indices\\
\hline
$\Upsilon$ & set of frequency bin values\\
\hline\hline 
\end{tabular}
\end{center}
\end{table}

\section{Preliminaries}\label{sec:preliminaries}

A graph $G\in\mathcal{G}(\mathcal{V},\mathcal{E})$ is a tuple, where $\mathcal{V}$ is the node set of cardinality $N$, and $\mathcal{E}$ is a set of pairs of nodes, referred to as edges. We assume all nodes are self-connected, i.e., for all $v\in \mathcal{V}$ the pair $(v,v)\in \mathcal{E}$. The adjacency matrix $A\in \lbrace 0,1\rbrace^{N\times N}$ is an alternate representation of $\mathcal{E}$:
\begin{align}\label{Adj_mat}
A_{u,v}=\left\{
  \begin{array}{l l}
    1  &  \quad \text{if  }\quad {(u,v)}\in \mathcal{E}  \\
    0  &  \quad \text{otherwise  }   \\
  \end{array} \right..
\end{align}

We define $\mathbf{x}_u\in\mathbb{R}^{T }$ as a signal (column) vector associated with node $u\in \mathcal{V}$, where $\mathbb{R}$ denotes the set of real numbers. A signal on the graph $G$ is then defined as: 
\begin{align}\label{x_def}
X=\left[\mathbf{x}_1 \quad \mathbf{x}_2 \quad \cdots \quad \mathbf{x}_N \right]^\top\in \mathbb{R}^{N\times T}.
\end{align}
A graph similarity matrix $S\in \mathbb{R}^{N\times N}$ is a function of the graph $G$, the graph signal, and model parameters $\params$:
\begin{align}
S = \func{f}{\textrm{XtoS}}(X, G;\params).
\end{align}
The value $S_{u,v}\in \mathbb{R}$ is the similarity between nodes $u,v \in \mathcal{V}$. The neighbourhood $\mathcal{N}_u$ of node $u$ is:
\begin{equation}
\mathcal{N}_u = \lbrace v\in \mathcal{V} | (u,v)\in \mathcal{E} \rbrace.
\end{equation}

We categorize the work on  graph learning into two main clusters: model-free and model-based. In model-free schemes, which include covariance and correlation-based methods, the graph is computed without enforcing any prior structure. In contrast, model-based methods assert a structure on the set of possible graphs. The assumptions in such models include: smoothness in which a signal changes smoothly between highly weighted connected nodes, e.g., \cite{egilmez2018graph}; diffusion which models the graph signal as a sum of heat diffusion processes, e.g., \cite{thanou2017learning}; and time-variability  in which both the spatial and temporal interdepencies in the signals are taken into account, e.g., \cite{shen2017kernel,linear_learning, shen2019nonlinear,SVARM_brain_connectivity}. Under the assumption of time-variability, linear and non-linear vector autoregressive models are used, the parameters of which describe the similarity matrix. Dynamic causal modeling falls into the last category \cite{friston2011functional}.

Model-free and model-based approaches provide a trade-off between computational requirements and richness in the ability to differentiate amongst various hypotheses. By imposing a prior structure on the graph signals, model-based methods are able to compare multiple hypotheses about interdependencies among nodes \cite{friston2011functional}. On the other hand, from a computational point of view,  existing model-based methods can be expensive. In the following we elaborate on the issue of computational inefficiency.

Consider $\tilde{I}$ samples of the graph signal $X$, forming the index set $\tilde{\mathcal{I}}$, where $X^{(i)}:i\in \tilde{\mathcal{I}}$, denotes the $i$th sample. Existing model-based  graph learning methods,  derive the similarity matrix of each sample by optimizing an  objective function $ \func{f}{\textrm{EC}}$. They can either use all the samples:
\begin{align}\label{EC_all}
\nonumber
&  \lbrace S^{(i)}: i\in \tilde{\mathcal{I}}\rbrace \\
& =\underset{\lbrace\tilde{S}^{(i)}: i\in \tilde{\mathcal{I}}\rbrace}{\textrm{argmin}} \func{f}{\textrm{EC}}(\lbrace X^{(i)}: i\in \tilde{\mathcal{I}}\rbrace;\lbrace\tilde{S}^{(i)}: i\in \tilde{\mathcal{I}}\rbrace),
\end{align}
or can infer similarity on a per sample basis:
\begin{align}\label{EC_per}
\begin{array}{c}
S^{(i)}= \underset{\tilde{S}^{(i)}}{\textrm{argmin}} \func{f}{\textrm{EC}}(X^{(i)};\tilde{S}^{(i)}).
\end{array}
\end{align}
We term such  graph learning methods ``edge-centric'' since they compute the similarity matrix $S^{(i)}$ by directly modeling pairwise connections.
The major drawback to these methods is their lack of scalability. If the topology of the graph $G$ or the signal $X^{(i)}$ defined on $G$ changes, the algorithm must be rerun from scratch to obtain $S^{(i)}$. Due to the large parameter space to be searched over, edge-centric schemes can be computationally intractable for large-scale problems. 

To address this issue, we introduce latent variables ($\params$) into our proposed model-based method, which we term ``node-centric'' (NC) graph learning. In this approach (to be explained later in detail), we use representation learning to obtain the similarity matrix. A high-level description is as follows. Representation learning introduces \textit{shared parameters} $\params^{\star}$ to transform the signal on the graph \textit{nodes} (this is why this method is termed ``node-centric''). The transformation finds embeddings for the nodes,
using the function $\func{f}{\textrm{XtoZ}}$:
\begin{align}\label{XtoZ}
Z = \func{f}{\textrm{XtoZ}}(X,A;\params^{\star}),
\end{align}
where
\begin{align}\label{Z}
Z=\left[\mathbf{z}_1 \quad \mathbf{z}_2 \quad \cdots \quad \mathbf{z}_N \right]^\top\in \mathbb{C}^{N\times D}.
\end{align}
We expand the feature domain to $\mathbb{C}$, the set of complex numbers, as complex numbers will be useful when dealing with the frequency domain in Sec.~\ref{subsec:initial_feature}. 
The similarity matrix is then computed from $Z$ using the function $\func{f}{\textrm{ZtoS}}$ as: 
\begin{align}\label{ZtoS}
S = \func{f}{\textrm{ZtoS}}(Z;\params^{\star}).
\end{align}

In brief, given $\params^{\star}$, the similarity matrix is obtained by applying the closed-form functions $\func{f}{\textrm{XtoZ}}$ and $\func{f}{\textrm{ZtoS}}$, both of which have low computational complexity. To reduce the computation required to find $\params^{\star}$ we assume the standard training and testing setup used in unsupervised machine learning. In a training phase, $I$ samples of $X$, forming the set of training indices $\mathcal{I}$, are used to optimize the $\params$ as
\begin{align}\label{param_optimize}
\params^{\star} = \underset{\params}{\textrm{argmin}} \func{f}{\textrm{NC}}(\lbrace X^{(i)}: i\in \mathcal{I}\rbrace,A;\params).
\end{align}
We assume the training and testing data have similar statistical properties and so, generalization of model parameters to new data is possible. As a result, the model parameters $\params^{\star}$ can be reused on test data to compute the similarity matrix using \eqref{XtoZ} and \eqref{ZtoS}. 
 

A simple way to quantify the claimed scalability of the node-centric method is to define computational complexity in terms of the number of variables to optimize. The \textit{online} computational complexity of the node-centric method, per testing sample, shrinks to $\mathcal{O}(1)$ from its edge-centric counterpart of  $\mathcal{O}(N^2)$ previously formulated in \eqref{EC_per}. The reason of such computational efficiency is that the shared parameters $\params^{\star}$ are reused in the testing phase, without the need to be re-optimized. 



\subsection{Representation Learning on Graphs}
Given a graph $G$, representation learning on the graph is a feature reduction technique that finds vector representations (embeddings) for graph nodes, by iteratively processing nodes' local neighbourhood information \cite{hamilton2017inductive}. Classical machine learning approaches (e.g., fully-connected and convolutional neural networks) allow inputs that are $1$D vectors or $2$D images. While such vectors and images are Euclidean data, graphs do not follow Euclidean geometry. As a result, representing a graph using feature vectors becomes important for graph-inspired analysis, such as graph classification \cite{hamilton2017representation}. 

Graph neural networks (GNNs) are one variant of graph representation learning. They have three properties that are relevant to the objective of this work. 
\begin{enumerate}
\item \textit{Parameter sharing}: Model parameters used to compute  the embeddings are shared among all nodes.
\item \textit{Inductive learning}: The model used to get the embeddings of the graph nodes can be generalized. In other words, after the model parameters are determined during the training phase, they can be reused to compute unseen nodes' embeddings in the testing phase \cite{mehler2018lure}. 
\item \textit{Feature-richness}: In early work on representation learning, graphs were feature-less. There was no signal assigned to the nodes. As a result, node embeddings were calculated only based on $G=(\mathcal{V},\mathcal{E})$~\cite{LINE,cao2015grarep,grover2016node2vec}. However, GNNs consider feature-rich graphs where a signal is initially assigned to each node. GNNs include the features of neighbouring nodes in the embedding computation.
\end{enumerate}

\subsection{GraphSAGE}\label{subsec:graphSAGE}
In this paper, we build on the GraphSAGE algorithm \cite{hamilton2017inductive}. In GraphSAGE, each node $u\in\mathcal{V}$ is assigned an initial feature vector $\mathbf{h}^0_u\in\mathbb{C}^{D_0 }$, the result of applying the function $q:\mathbb{R}^{T } \rightarrow \mathbb{C}^{D_0 }$ to the node's signal $\mathbf{x}_u$: 
\begin{align}\label{q_def}
\mathbf{h}^0_u=q(\mathbf{x}_u).
\end{align}
The goal of GraphSAGE is to find an embedding for each node $u$:
\begin{align}
\mathbf{z}_u=\func{f}{\textrm{EMB}}(u,X,A;\Psi),
\end{align}
 
which yields the general function \eqref{XtoZ} after stacking up all the per node embeddings $\mathbf{z}_u$ into a matrix. In the following paragraphs, we will elaborate on the steps through which $\func{f}{\textrm{EMB}}$ is applied to the graph signal to yield the final embeddings.
 
The vector $\mathbf{z}_u$, previously introduced in \eqref{Z}, is defined in a way that it incorporates information on the graph topology. Graph topology information includes the graph community (group of internally densely connected nodes) that the nodes belongs to, and types of the nodes in the graph \cite{hamilton2017representation} (e.g., isolated, leaf, bridge \cite{rossi2014role}).

GraphSAGE iteratively updates its features $K$ times. At iteration $k\in[K]\triangleq \lbrace 1,2,\cdots,K\rbrace$, a \textit{hidden} feature vector $\mathbf{h}_u^k\in \mathbb{C}^{D_0 }$ is computed using the hidden features of node $u$ and its neighbours from the previous iteration. To accomplish this, an aggregation function $g_u^k:\mathbb{C}^{|\mathcal{N}_u|\times D_0}\rightarrow \mathbb{C}^{D_0}$, where $|.|$ is the cardinality operator, aggregates the features of node $u$'s neighbours to produce:
\begin{align}\label{g_def}
\mathbf{h}_u^k = g_u^k\left(\left\lbrace \mathbf{h}_v^{k-1} : v\in \mathcal{N}_u\right\rbrace ; \Psi^k\right).
\end{align}
In \eqref{g_def}, $\Psi^k$ is the set of model parameters at iteration $k$. Various functions $g_u^k$ are used in \cite{hamilton2017inductive}, among which we consider $g_u^{k,\textrm{mean}}$ and $g_u^{k,\textrm{max}}$, parameterized by $\Psi^k=\lbrace {U}^{k}, \mathbf{b}^{k}\rbrace$. The parameters ${U}^{k}\in \mathbb{R}^{D_0\times D_0},\mathbf{b}^{k}\in \mathbb{R}^{D_0 }$ specify different choices of $g_u^k$:
\begin{align}\label{g_mean}
\nonumber
&g_u^{k,\textrm{mean}} \left(\left\lbrace \mathbf{h}_v^{k-1} : v\in \mathcal{N}_u\right\rbrace;{U}^{k},\mathbf{b}^{k}\right) \\
& = \sigma\left(\frac{1}{{|\mathcal{N}_u|}} {U}^{k}{\displaystyle\sum_{v \in \mathcal{N}_u }\mathbf{h}^{k-1}_v}+\mathbf{b}^{k}\right),
\end{align}
\begin{align}\label{g_max}
\nonumber
&g_u^{k,\textrm{max}} (\left\lbrace \mathbf{h}_v^{k-1} : v\in \mathcal{N}_u\right\rbrace;{U}^{k}, \mathbf{b}^{k}) \\
&= \max\left\lbrace\sigma\left( {U}^{k}{\mathbf{h}^{k-1}_v}+\mathbf{b}^{k}\right): v \in \mathcal{N}_u \right\rbrace,
\end{align}
where $\sigma$ is an activation function that introduces non-linearity into the aggregation functions. In this work, two choices for $\sigma$, namely the rectifier linear unit (\textit{ReLU}) and the \textit{softmax} functions, are implemented \cite{nwankpa2018activation}. 


The concatenation of each node's initial and the final hidden feature vectors is the node's embedding: 
\begin{align}\label{z_def}
\mathbf{z}_u= \left[\begin{array}{l l l}\mathbf{h}^0_u \\ \mathbf{h}_u^K\end{array}\right] \in \mathbb{C}^{D },
\end{align} 
where $D=2D_0$. 
 
\begin{remark}
Later when we define the similarity matrix, we will see how the embedding vectors $\mathbf{z}_u$ are weighted and transformed to construct $S$. When we include both the initial features and the final hidden feature vectors in  the definition of $\mathbf{z}_u$ in \eqref{z_def}, we assure retrieving conventional similarity matrix, which uses only the initial features, is always possible. Such recovery is done by setting the weights corresponding to the hidden features to zero. Therefore, the concatenation guarantees that $S$ is a generalized version of conventional methods which is rooted in assigning different weights to the elements of $\mathbf{z}_u$. 
\end{remark}
 
\begin{figure}[!t]
\centering
\includegraphics[width=3.5in]{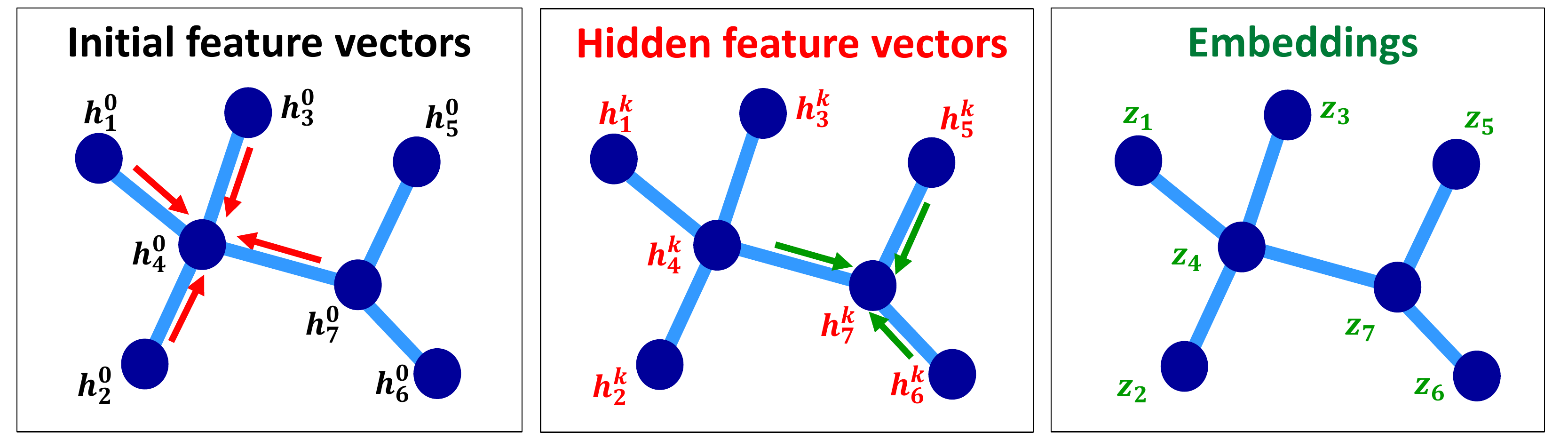}
\caption{Feature aggregation process in GraphSAGE \cite{hamilton2017inductive}.}
\label{fig.feature_aggregation}
\end{figure}

Fig.~\ref{fig.feature_aggregation} depicts an overview of the aggregation process employed to generate node embeddings.

In order to determine the parameters $\Psi^k,k\in[K]$ used in $g_u^k$ in \eqref{g_def}, we intend to specify the optimization of \eqref{param_optimize}. We follow the guidelines in the original representation learning formulation of \cite{LINE}, rather than the approximated version used in the GraphSAGE paper \cite{hamilton2017inductive}. First, a conditional probability distribution $\mathbf{p}^{\textrm{gSAGE}}(.|v;Z)$ is defined by softmaxing the inner product of the pairwise embeddings as \cite{LINE,cao2015grarep,grover2016node2vec}:
\begin{align}\label{p_gsage}
\mathbf{p}^{\textrm{gSAGE}}(u|v;Z)=\frac{\exp(\mathbf{z}_u^\top \mathbf{z}_v)}{\displaystyle\sum_{\tilde{u} \in \mathcal{V}} \exp(\mathbf{z}_{\tilde{u}}^\top \mathbf{z}_v)}.
\end{align}
Note that \eqref{p_gsage} is also parameterized by $\Psi^k,k\in[K]$, due to its use in the definition of $Z$ (cf. \eqref{g_def} and \eqref{z_def}). Nevertheless, $\Psi^k,k\in[K]$ is dropped from notations for simplicity. Based on the graph topology, we define the target probability distribution $\hat{\mathbf{p}}(.|v)$ as:
\begin{align}\label{p_hat}
\hat{\mathbf{p}}(u|v) = \left\{
  \begin{array}{l l}
    \frac{1}{|\mathcal{N}_v|}  &  \quad \text{if  }\quad {(u,v)}\in \mathcal{E}  \\
    0  &  \quad \text{otherwise  }   \\
  \end{array} \right..
\end{align}
Graph-based similarity among the embeddings is achieved by minimizing the Kullback–Leibler (KL) divergence between $\mathbf{p}^{\textrm{gSAGE}}(.|v;Z)$ and $\hat{\mathbf{p}}(.|v)$:
\begin{align}\label{KL_min}
\begin{array}{l l l}
\displaystyle\min_{\left\lbrace \Psi^k: k\in[K]\right\rbrace}  & \displaystyle\sum_{v\in \mathcal{V}} \lambda_v\textrm{KL}\left(\hat{\mathbf{p}}(.|v)\Vert\mathbf{p}^{\textrm{gSAGE}}(.|v;Z)\right).
\end{array}
\end{align}
By expanding the expressions in \eqref{KL_min} and by substituting from \eqref{p_hat}, where for simplicity we let $\lambda_v=|\mathcal{N}_v|$ as in \cite{LINE}, it is straightforward to get:
\begin{align}\label{gSAGEsteps}
\begin{array}{l l}
\displaystyle\sum_{v\in \mathcal{V}} \lambda_v\textrm{KL}\left(\hat{\mathbf{p}}(.|v)\Vert\mathbf{p}^{\textrm{gSAGE}}(.|v;Z)\right) \\
\qquad = \displaystyle\sum_{v\in \mathcal{V}} \displaystyle\sum_{u\in \mathcal{V}} \lambda_v\hat{\mathbf{p}}(u|v) \log\left(\frac{\hat{\mathbf{p}}(u|v)}{\mathbf{p}^{\textrm{gSAGE}}(u|v;Z)}\right)\\
\qquad = -\displaystyle\sum_{v\in \mathcal{V}} \displaystyle\sum_{u\in \mathcal{N}_v} \log\left({\mathbf{p}^{\textrm{gSAGE}}(u|v;Z)}\right)+\log(|\mathcal{N}_v|).
\end{array}
\end{align}
By taking out the constants in \eqref{gSAGEsteps}, the optimization problem in \eqref{KL_min} can be rewritten as:
\begin{align}\label{gSAGE}
\begin{array}{l l l}
\displaystyle\min_{\left\lbrace \Psi^k:k\in[K]\right\rbrace}  & -\displaystyle\sum_{(u,v)\in \mathcal{E}} \log(\mathbf{p}^{\textrm{gSAGE}}(u|v;Z)).
\end{array}
\end{align}

\section{Proposed Data-Driven Graph Learning}\label{sec:problem_statement}
 
In this section, we modify the formulation \eqref{gSAGE} to address the intended problem and applications of this work. To do so, we first talk about the assumptions of GNNs. Next, we outline two objectives of this work that differ from those of GNNs. We then approach each objective through a modification to the GNN's optimization problem. Each subsection is devoted to detailed explanations on how each modification is applied.

The underlying assumptions of GNNs are twofold. First, there is \textit{one massive graph}. Second, the goal is to compute an embedding for each graph node. The first assumption is not directly applicable to cases involving multiple graphs. For instance, time-variable signals in brain applications lead to the construction of multiple time-varying graphs \cite{sizemore2018dynamic}.\footnote{ There have been some efforts in the literature to extract one large adjacency matrix from time-varying signals using the concept of spatio-temporal (multilayer) graphs \cite{bassett2013robust,pedersen2018multilayer,ghoroghchian2018hierarchical}. However, the suitability of using such graphs as the basis of a GNN are questionable due to their exponentially growing size.  }
Additionally, the second assumption is not in line with the goal of this work which is learning the graph similarity matrix. As a result, we propose two modifications to GNNs to address the two distinguished objectives of this work. 

The first objective is to shift the goal of learning from learning the nodes' embeddings in GNNs to learning the similarity matrix. This shift necessitates a number of changes to  GraphSAGE when the specific application of brain network modeling comes into play. The changes include the optimization problem formulation, the initial feature computation, and the definition of the similarity matrix and parameters.

\begin{figure}[!t]
\centering
\subfloat[Inter-graph generalization in representation learning on graphs]{\includegraphics[width=2in]{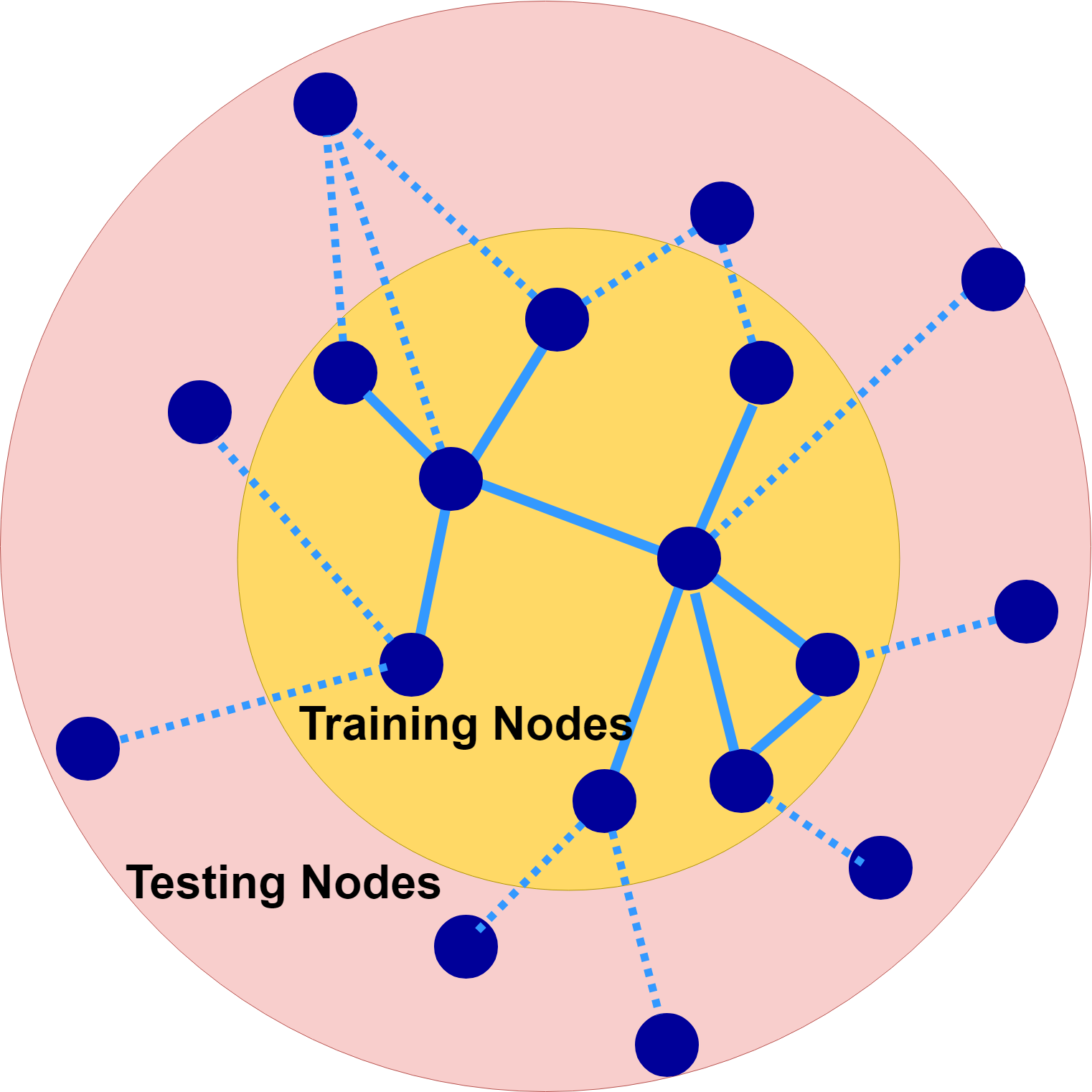}
\label{fig.InterGraph}}
\hfil
\subfloat[Intra-graph generalization in this paper]{\includegraphics[width=2.5in]{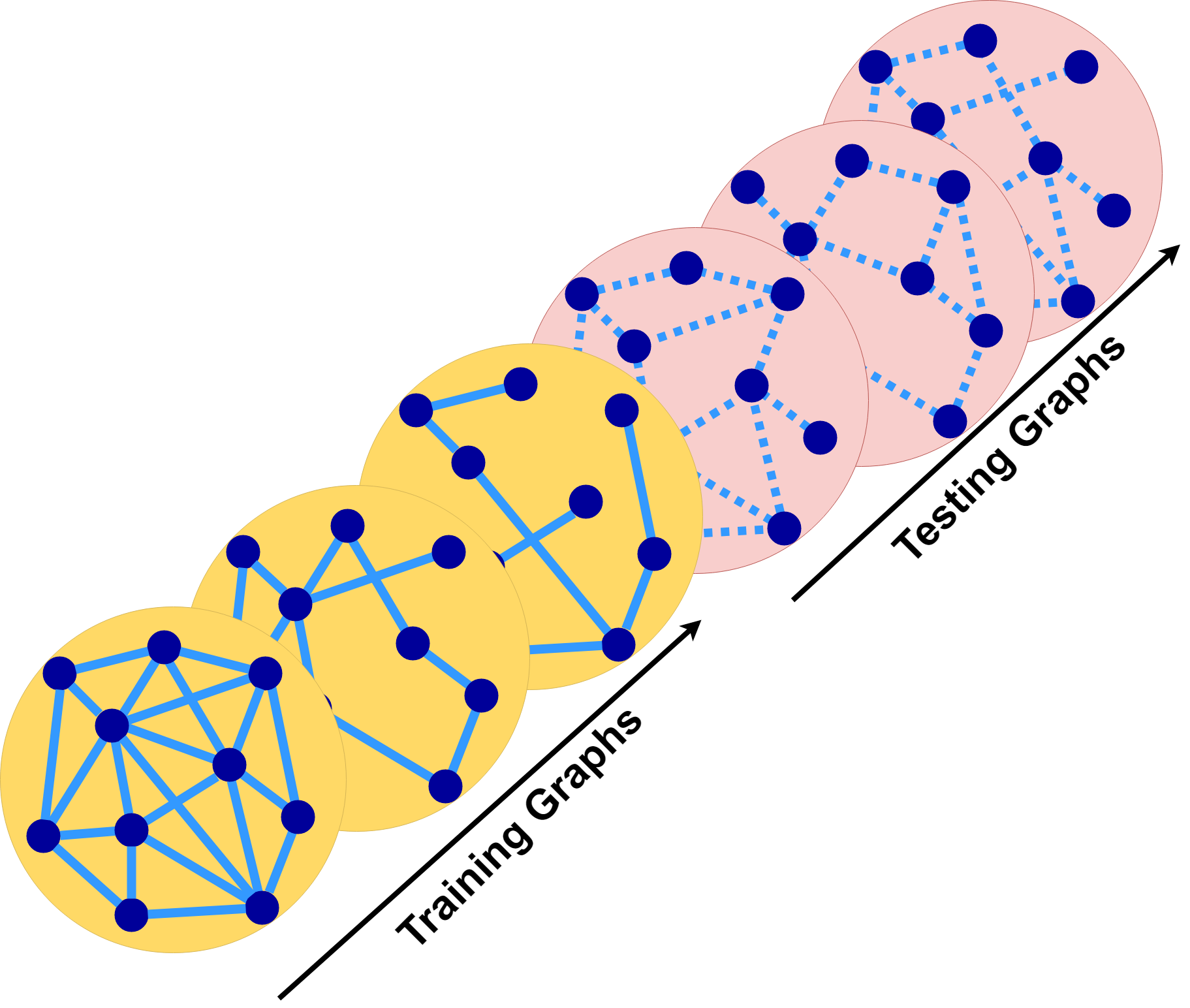}%
\label{fig.IntraGraph}}
\caption{Two generalization techniques for the estimation of the similarity matrices }
\label{fig.Inter_IntraGraph}
\end{figure}

Our second objective is  to learn \textit{many small graphs}. In GNNs, in order to learn (optimize) the embeddings' parameters $\Psi^k,k\in[K]$ in \eqref{g_def}, the graph nodes are split into two groups as shown in Fig.~\ref{fig.InterGraph}, i.e. into training and testing nodes. In the training phase, the training nodes are used to learn  the parameters. In the testing phase, the learned parameters are used to compute the embeddings of the testing nodes. Such a train-test splitting procedure is a way to achieve a generalization, which we call \textit{inter-graph generalization}. By changing the learning target from one graph to many graphs, we modify the train-test splitting technique as depicted in Fig.~\ref{fig.IntraGraph}; all nodes of a number of graphs are used for training, while the rest of the graphs are used for testing. We call this approach \textit{intra-graph generalization}.

In the following subsections we elaborate on these two modifications we made to GraphSAGE. We derive an optimization problem formulation for our proposed node-centric data-driven (NCDD) graph learning method. 

\subsection{First Modification: Learning the Similarity Matrix}
As explained before, the formulation in \eqref{gSAGE} enforces graph-based similarity among the embeddings. We first define the similarity matrix as a function $\func{f}{\textrm{NCDD}}$ of the embeddings, parametrized by $\mathbf{\Theta}$:
\begin{align}\label{Similarity_pairwise}
S_{u,v} = \func{f}{\textrm{NCDD}}(\mathbf{z}_u, \mathbf{z}_v;\mathbf{\Theta}).
\end{align}  
 
It should be noted that \eqref{Similarity_pairwise} is an elementwise version of the previously introduced general function \eqref{ZtoS}. The exact definitions of $\func{f}{\textrm{NCDD}}$ and the parameter set $\mathbf{\Theta}$, which gives different weights to each embedding component, will be later introduced in Sec.~\ref{subsec:similarity_mat}.

Next, we change the conditional probability $\mathbf{p}^{\textrm{gSAGE}}$ in \eqref{p_gsage} to $\mathbf{p}^{\textrm{NCDD}}$ as:
\begin{align}\label{similarity_measure_softmax}
\nonumber
&\mathbf{p}^{\textrm{NCDD}}(u|v;Z, \mathbf{\Theta}) =\frac{\exp(\func{f}{\textrm{NCDD}}(\mathbf{z}_u, \mathbf{z}_v;\mathbf{\Theta}))}{\displaystyle\sum_{\tilde{u} \in \mathcal{V}} \exp(\func{f}{\textrm{NCDD}}(\mathbf{z}_{\tilde{u}}, \mathbf{z}_v;\mathbf{\Theta}))}\\
&\qquad\qquad\qquad\qquad\quad=\frac{\exp(S_{u,v})}{\displaystyle\sum_{\tilde{u} \in \mathcal{V}} \exp(S_{\tilde{u},v})}.
\end{align}
In the next section, two definitions for $\func{f}{\textrm{NCDD}}$ are provided based on two domains, time and frequency. By replacing $\mathbf{p}^{\textrm{gSAGE}}$ with $\mathbf{p}^{\textrm{NCDD}}$, an optimization problem similar to \eqref{gSAGE} is derived:
\begin{equation}\label{repL_new}
  \begin{array}{l l l}
    \displaystyle\min_{\left\lbrace \Psi^k:k\in[K]\right\rbrace, \mathbf{\Theta}} &  -\displaystyle\sum_{(u,v)\in \mathcal{E}} \log\left(\mathbf{p}^{\textrm{NCDD}}(u|v;Z,\mathbf{\Theta})\right) \end{array}
\end{equation}
where for all $k\in [K]$ and for all $u\in\mathcal{V}$:
\begin{align}
\begin{array}{l l l}
 \mathbf{z}_u = \left[\begin{array}{l l l}\mathbf{h}^0_u \\ \mathbf{h}_u^K\end{array}\right],\\
 \mathbf{h}_u^{k} =  g\left(\left\lbrace \mathbf{h}_v^{k-1} : v\in \mathcal{N}_u \right\rbrace ; \Psi^k\right),\\
\mathbf{h}^0_u=q(\mathbf{x}_u).\\
 \end{array}
\end{align}
Using \eqref{similarity_measure_softmax}, the objective function \eqref{repL_new} is rewritten as:
\begin{align}\label{repL_new_simplified}
\begin{array}{l l}
& -\displaystyle\sum_{(u,v)\in \mathcal{E}} \log\left(\mathbf{p}^{\textrm{NCDD}}(u|v;Z,\mathbf{\Theta})\right)\\
&= -\displaystyle\sum_{(u,v)\in \mathcal{E}} \left[S_{u,v}- \log \displaystyle\sum_{\tilde{u}\in \mathcal{V}}\exp(S_{\tilde{u},v})\right]\\
&=-\displaystyle\sum_{v\in \mathcal{V}}\left[\displaystyle\sum_{u\in \mathcal{N}_v} S_{u,v} - |\mathcal{N}_v| \log \displaystyle\sum_{\tilde{u}\in \mathcal{V}}\exp(S_{\tilde{u},v})\right].
\end{array}
\end{align}

\subsection{Second Modification: Intra-graph Generalization}
The transition from inter-graph generalization to intra-graph generalization calls for another modification to the optimization problem in \eqref{repL_new}. Reusing the notations of Sec.~\ref{sec:preliminaries}, consider $I$ samples of the signal on graph, denoting sample $i\in\mathcal{I}$ by $X^{(i)}$. We reserve the superscript $i$ for indexing the feature vectors and the similarity matrices corresponding to the $i$th sample. We assume the underlying adjacency matrix $A$ and model parameters $\left\lbrace \Psi^k:k\in[K]\right\rbrace,\mathbf{\Theta}$ are fixed for all samples. By incorporating the samples, we change the optimization objective in \eqref{KL_min} into:
\begin{align}\label{KL_min_new}
\begin{array}{l l l}
\displaystyle\sum_{i\in\mathcal{I}}\displaystyle\sum_{v\in \mathcal{V}} \lambda_v\textrm{KL}\left(\hat{\mathbf{p}}(.|v) \Vert \mathbf{p}^{\textrm{NCDD}}(.|v;Z^{(i)},\mathbf{\Theta})\right).
\end{array}
\end{align}
Following similar steps as in \eqref{gSAGEsteps}, the objective in \eqref{KL_min_new} simplifies to: 
\begin{align}\label{representation learning_obj_samples}
\begin{array}{l l l}
 -\displaystyle\sum_{i\in\mathcal{I}}\displaystyle\displaystyle\sum_{(u,v)\in \mathcal{E}} \log\left(\mathbf{p}^{\textrm{NCDD}}(u|v;Z^{(i)},\mathbf{\Theta})\right).
\end{array}
\end{align}
Finally,  by indexing $S^{(i)}_{u,v} = \func{f}{\textrm{NCDD}}(\mathbf{z}^{(i)}_u, \mathbf{z}^{(i)}_v;\mathbf{\Theta})$ and substituting \eqref{repL_new_simplified} into \eqref{representation learning_obj_samples}, the final NCDD  graph learning optimization problem is derived:
\begin{align}\label{problem_statement}
   \displaystyle\min_{\left\lbrace \Psi^k:k\in[K]\right\rbrace, \mathbf{\Theta}}   -\displaystyle\sum_{i\in\mathcal{I}}\displaystyle\sum_{v\in \mathcal{V}}\left[\displaystyle\sum_{u\in \mathcal{N}_v}S_{u,v}^{(i)} -
    |\mathcal{N}_v| \log \displaystyle\sum_{\tilde{u}\in \mathcal{V}}\exp(S_{\tilde{u},v}^{(i)})\right],
\end{align}
where for all $i\in\mathcal{I}$, for all $k\in [K]$, and for all $u\in\mathcal{V}$
\begin{equation}
\begin{array}{l l l}
S^{(i)}_{u,v} =\func{f}{\textrm{NCDD}}(\mathbf{z}^{(i)}_u, \mathbf{z}^{(i)}_v;\mathbf{\Theta}),\\
 \mathbf{z}^{(i)}_u = \left[\begin{array}{l l l}\mathbf{h}^{0,(i)}_u, \\ \mathbf{h}^{K,(i)}_u\end{array}\right], \\
\mathbf{h}^{k,(i)}_u =  g\left(\left\lbrace \mathbf{h}^{k-1,(i)}_v:v\in \mathcal{N}_u \right\rbrace ; \Psi^k\right),\\
\mathbf{h}^{0,(i)}_u=q(\mathbf{x}^{(i)}_u).\\
 \end{array} 
\end{equation}

\section{Brain Connectivity Inference}\label{sec:brain_connectivity}
In this section, we apply the NCDD  graph learning method to compute brain connectivity. We define the brain connectivity as the similarity matrix $S$ computed from the brain signals. We present tools to construct $S$ based on the time and frequency domain signals, which are respectively linked to the notions of correlation and coherence. Although NCDD is a model-based method, since it statistically measures the interrelationships between neuronal events, it falls into the category of brain functional connectivity. The computations provided in this section are identical for all samples. So, to simplify notation, we drop the superscript $i$ in this section and denote a sample of the graph signal by $X$.

We refer to an arbitrary vector and matrix as $\genvec\in\mathbb{C}^M$ and $\genmat\in\mathbb{C}^{M\times L}$, respectively. We can substitute them with different vectors or matrices, e.g. with the hidden feature vector $\mathbf{h}_u^k$. We introduce $[\genvec]_{\mathcal{M}}$ and $[\genmat]_{\mathcal{M},\mathcal{L}}$ as the notations for indexing, i.e., they point out to sets of components corresponding to indices $\mathcal{M}\subseteq[M]$ and $\mathcal{L} \subseteq [L]$, respectively.

In the context of brain signal processing, $\mathbf{x}_u$ represents the time-series acquired from electrode (channel) $u$ over $T$ units of time. Next, we consider two forms of the  signal-to-initial-feature mapping $q$, from which the initial node features are computed. Each of the two forms correspond to the time and frequency domains and are linked to two conventional brain network modelings. 

\subsection{Initial Feature Computation}\label{subsec:initial_feature}
In the time-domain (TD) analysis, we define $q$ in \eqref{q_def} as an identity function, i.e., the initial feature vectors are set equal to the graph signal $\mathbf{h}^{0,\textrm{TD}}_u=\mathbf{x}_u$, where $ u\in \mathcal{V}, \mathbf{x}_u\in\mathbb{R}^{D_0}$ and $D_0 = T$. Through these choices, the resulting similarity matrix will be related to the notion of correlation (to be discussed in Sec. \ref{subsec:similarity_mat}). 

In the frequency domain, we define the function $q$ of \eqref{q_def} as a series of computations from the graph signal $X$ to the initial feature vectors. We perform the computation process in such a way that the ensuing similarity matrix is linked to the notion of coherence (to be explained in Sec. \ref{subsec:similarity_mat}). The process is as follows. Initially, we partition each node's sample into $\tilde{T}$ inner windows, each of size $W$, using the function $\func{f}{\textrm{win}}:\mathbb{R}^{N\times T}\rightarrow\mathbb{R}^{N \times \tilde{T}\times W }$:
\begin{align}
\tilde{X}=\func{f}{\textrm{win}}(X).
\end{align}
Next, we calculate the Discrete Fourier Transform (DFT) of $\tilde{X}$, for each channel $v\in \mathcal{V}$ and each inner window $t\in [\tilde{T}]$, over $W$ frequency bins using the function $\func{f}{\textrm{DFT}}: \mathbb{R}^{N \times \tilde{T}\times W }\rightarrow \mathbb{C}^{N \times \tilde{T}\times W }$:
\begin{align}\label{DFT_def}
\arc{X}=\func{f}{\textrm{DFT}}(\tilde{X}).
\end{align}


Let $\mathcal{W}=[W]$ be the set of frequency bin indices, $\Upsilon\in\mathbb{R}^{W}$ denote the set of frequency bin values, and $\tilde{\mathcal{T}}=[\tilde{T}]$ be the set of all inner window indices. The two-dimensional matrix $[\arc{X}]_{v, \tilde{\mathcal{T}}, \mathcal{W}}$ corresponding to the $v$th component is vectorized using the $\textrm{vec}:\mathbb{C}^{\tilde{T}\times W }\rightarrow \mathbb{C}^{\tilde{T}W }$ operator which concatenates the input's columns. The initial frequency domain (FD) feature vector is:
\begin{align}\label{vectorizing}
\mathbf{h}^{0,\textrm{FD}}_v=\textrm{vec}([\arc{X}]_{v, \tilde{\mathcal{T}}, \mathcal{W}})\in\mathbf{C}^{\tilde{T}W},
\end{align}
where $D_0=\tilde{T}W$.
\subsection{Similarity Matrix Definition}\label{subsec:similarity_mat}
In the time-domain analysis, we define the similarity matrix \eqref{Similarity_pairwise} as a weighted version of correlation. To do so, we first denote a centering-normalizing (CN) operator by $\func{f}{\textrm{CN}}:\mathbb{C}^{D }\rightarrow \mathbb{C}^{D }$, where $d\in [D]=[2D_0]$:
\begin{align}
[\func{f}{\textrm{CN}}(\genvec)]_{d}=\frac{([\genvec]_{d}-\overline{\genvec})}{\sqrt{\frac{1}{D-1}\displaystyle\sum_{d\in [D]} ([\genvec]_{d}-\overline{\genvec})^2}},
\end{align}
and $\overline{\genscalar}=\frac{1}{D}\displaystyle\sum_{d\in [D]} [\genvec]_d$ is the average of input.  Also, the output of $\func{f}{\textrm{diag}}: \mathbb{R}^{D }\rightarrow \mathbb{R}^{D\times D}$ is an all-zero matrix except for the diagonal:
\begin{align}
    \func{f}{\textrm{diag}}\left(\genvec\right)
    =\left[\begin{array}{cccc}
    [\genvec]_1 &  & &\mathbf{0}\\
     & [\genvec]_2 & &\\
     & & \ddots &\\
     \mathbf{0}&&&[\genvec]_D\\
    \end{array}\right].
\end{align}
Next, $\func{f}{\textrm{NCDD}}$ of \eqref{Similarity_pairwise} in the time domain (TD) is defined as a weighted inner product between pairwise node embeddings, using the parameter set defined as $\mathbf{\Theta} =\left\lbrace \mathbf{\theta}\in \mathbb{R}^{D }\right\rbrace$:
\begin{align}
S_{u,v}=\func{f}{\textrm{NCDD}}(\mathbf{z}_u, \mathbf{z}_v;\mathbf{\Theta}) = \func{f}{\textrm{CN}}(\mathbf{z}_u)^\top  \func{f}{\textrm{diag}}(\mathbf{\theta}) \func{f}{\textrm{CN}}(\mathbf{z}_v).
\end{align}

In the frequency-domain analysis, we first reshape the vectorized embeddings into a tensor;  we apply the inverse of the $\textrm{vec}$ operator \eqref{vectorizing} to each of the two parts of a node's embedding \eqref{z_def}, separately:
\begin{align}
\begin{array}{l}
\arc{\mathbf{z}}_u^{a}= \textrm{vec}^{-1}([\mathbf{z}_u]_{[D_0]})\in \mathbb{C}^{ \tilde{T}\times W}\\
\arc{\mathbf{z}}_u^{b}= \textrm{vec}^{-1}([\mathbf{z}_u]_{\lbrace D_0+1, D_0+2, \cdots, 2D_0= D\rbrace})\in \mathbb{C}^{ \tilde{T}\times W}
\end{array}.
\end{align}
From now on, we use $\twoparts\in\lbrace a,b \rbrace$  to refer to either of the parts, $a$ or $b$. Let $\arc{Z}^\twoparts\in \mathbb{C}^{ N\times \tilde{T}\times W}$ be the concatenation of $\arc{\mathbf{z}}_u^{a}: \forall u\in \mathcal{V}$ in a three-dimensional tensor. From the embedding definition in \eqref{z_def}, we have $\arc{Z}^{a}=\arc{X}$. Furthermore, ${\arc{Z}^{b}}$ is the final hidden feature vector, where $\arc{X}$ of \eqref{DFT_def} is the initial feature vector in the frequency domain. Hence, we treat ${\arc{Z}^{b}}$ as a post-processed DFT of the signal. Next, we use the cross-spectrum formula, which is the non-normalized version of coherence, to define the similarity matrix $S$ using the embeddings in the frequency domain. 
 
We use Welch's method to approximate the cross-spectrum \cite{kramer2013introduction}. In this method, we temporally divide the signal in each channel into inner windows. The cross spectrum is stored in a three-dimensional tensor $\Omega^{\twoparts}\in \mathbb{R}^{N \times N\times W}$. The cross spectrum between nodes $u,v$ in a frequency bin index $\omega$ is defined as the following inner product: 
\begin{equation}\label{welch_tensor}
\Omega^{\twoparts}_{u,v,\omega}=
\left|\sum_{t\in \tilde{\mathcal{T}}}\matidx{\arc{Z}^{\twoparts}}{u,t,\omega}\matidx{\arc{Z}^{\twoparts}}{v,t,\omega}^{*}\right|,
\end{equation}
where $(.)^*$ denotes complex conjugate. We can rewrite the component-wise definition of $\Omega^{\twoparts}$ in \eqref{welch_tensor} in a compact form. For this purpose, we first rearrange the components of $\arc{Z}^\twoparts \in\mathbb{C}^{N\times \tilde{T}\times W}$ in a two-dimensional matrix: 
\begin{align}
\begin{array}{l l l}
\mathcal{Z}^{\twoparts} &= \left[\begin{array}{l l l l}
\matidx{\arc{Z}^{\twoparts}}{\mathcal{V},1,1} & \matidx{\arc{Z}^{\twoparts}}{\mathcal{V},2,1} & . & \matidx{\arc{Z}^{\twoparts}}{\mathcal{V},\tilde{T},1}\\
\matidx{\arc{Z}^{\twoparts}}{\mathcal{V},1,2} & \matidx{\arc{Z}^{\twoparts}}{\mathcal{V},2,2} & . & \matidx{\arc{Z}^{\twoparts}}{\mathcal{V},\tilde{T},2}\\
\cdots & \cdots & . &\cdots\\
\matidx{\arc{Z}^{\twoparts}}{\mathcal{V},1,W} & \matidx{\arc{Z}^{\twoparts}}{\mathcal{V},2,W} & . & \matidx{\arc{Z}^{\twoparts}}{\mathcal{V},\tilde{T},W}\\
\end{array}\right]
\end{array}.
\end{align}
Then, $\Omega^{\twoparts}$ in \eqref{welch_tensor} is written in the following tensor form:
\begin{align}
\Omega^{\twoparts} =\func{f}{\textrm{blkdg}}(|\mathcal{Z}^{\twoparts} (\mathcal{Z}^{\twoparts})^\mathrm{H}|),
\end{align}
where $(.)^\mathrm{H}$ denotes matrix conjugate transpose, and $\func{f}{\textrm{blkdg}}:\mathbb{R}^{NW \times NW}\rightarrow \mathbb{R}^{N\times N\times W}$ is the block diagonal operator. The operator $\func{f}{\textrm{blkdg}}$ outputs $W$ blocks of size $N \times N$, on the diagonal of the input $|\mathcal{Z}^{\twoparts} (\mathcal{Z}^{\twoparts})^\mathrm{H}|$. Finally, the similarity matrix $S$ is defined as a weighted sum of $\matidx{\Omega^{\twoparts}}{\mathcal{V},\mathcal{V},\omega}:\forall \omega\in\mathcal{W}$. The weighting is determined by the set of parameters $\mathbf{\Theta}=\lbrace\mathbf{\theta}^a,\mathbf{\theta}^b\rbrace$, where $ \mathbf{\theta}^\twoparts\in \mathbb{R}^{ W}$ gives different weights to each frequency bin:
\begin{align}
S = \displaystyle\sum_{\twoparts\in\lbrace a,b \rbrace,\omega\in\mathcal{W}} \matidx{\theta}{\omega}^\twoparts \matidx{\Omega^{\twoparts}}{\mathcal{V},\mathcal{V},\omega}.
\end{align}


\subsection{Graph Topology} 
Thus far, our method based on representation learning assumed the graph topology $\mathcal{E}$, or equivalently $A$, is known. This assumption does not hold in many applications, including brain modeling using intracranial electroencephalographic (iEEG) data \footnote{ 
iEEG is but one method for estimating macroscale brain networks. Magnetoencephalography (MEG), functional magnetic resonance imaging (fMRI), and diffusion tensor imaging (DTI) are also commonly used \cite{bullmore2009complex}. We use iEEG here as it possible to record iEEG data for much longer periods of time in a single individual (e.g., days) than other methods, though it does not provide the comprehensive brain coverage afforded by DTI and fMRI.
 }. 

We estimate $A$ from the iEEG data. Such data-driven graph topology inference is necessary when using macroscale neural data such as iEEG because the spatial resolution of the data is too crude to be able to map the precise anatomical connections between the cells generating the iEEG signal. For example, the iEEG signal is dominated by cells within about $3$ mm of the electrode contacts \cite{dubey2019cortical}, an area of tissue that contains around $0.5$ million neurons \cite{Valiante_2012}. The area could have diverse patterns of anatomical connectivity. An  alternative approach is to obtain diffusion tensor imaging (DTI) information.   DTI is the closest approximation of $A$ from current sources of imaging and neural data, which is often used to estimate macroscale structural connectivity in human brains \cite{bullmore2009complex}.   However, such information is not available in the EU human iEEG epilepsy database \cite{EU_dataset} that is used in this work. Future work would involve obtaining $A$ from DTI to evaluate the accuracy of the data-driven approach we present to derive $A$.


In the following, the approach we develop to infer $A$ from the data is explained. Assuming a multivariate Gaussian distribution for the $i$th sample of the graph signal, the sample covariance matrix $P^{(i)}\in\mathbb{R}^{N\times N}$ is defined as:
\begin{align}\label{sample_cov_mat}
P^{(i)} = \frac{1}{D_0-1}(X^{(i)}-\overline{X}^{(i)}\mathbf{1}_{N})(X^{(i)}-\overline{X}^{(i)}\mathbf{1}_{N})^\top.
\end{align}
In \eqref{sample_cov_mat}, $\mathbf{1}_{N}$ is an all-one row vector of length $N$ and $\overline{X}^{(i)}\in \mathbb{R}^{N }$ is the average of the signal on each node: 
\begin{align}
\overline{X}^{(i)}=\frac{1}{D_0}\displaystyle\sum_{d=1}^{D_0} \matidx{X^{(i)}}{\mathcal{V},d}.
\end{align}
The inverse of the sample covariance matrix reveals direct (rather than indirect) dependencies in a graph. In other words, an inverse covariance component $\matidx{(P^{(i)})^{-1}}{u,v}$, conditions independence between the signal of node $u$ from the signal of node $v$, conditional on the signals of all other nodes. This feature makes the inverse covariance a good data-driven choice to infer the graph topology \cite{loh2012structure}. As a result, we define the adjacency matrix as follows, where $\eta$ is a binarizing threshold:
\begin{align}\label{graph_topology}
A_{u,v} = \left\{
  \begin{array}{l l}
    1  &  \quad \text{if  }\quad \frac{1}{I}\displaystyle\sum_{i\in\mathcal{I}} \matidx{(P^{(i)})^{-1}}{u,v}\geq \eta  \\
    0  &  \quad \text{otherwise  }   \\
  \end{array} \right..
\end{align} 
Note that the primary condition for the inverse matrices $(P^{(i)})^{-1}$ to be invertible, is that the signal dimension $D_0$ must be greater than the number of nodes $N$.

\section{Experiments}\label{sec:results}
 
In this section, we use the NCDD graph learning method to infer time-varying brain networks. As explained before, one of the uses of the network perspective to brain modeling is to distinguish different brain states. Our graph learning method can be used for various brain state identification purposes. Examples include the classification of sleep vs. non-sleep, poor memory function vs. good function, seizure vs. non-seizure. In this paper, we present results on seizure detection. We further aim to identify brain state in an \textit{online} manner. In other words, we develop an algorithm that constructs graphs in real time and estimates the corresponding state of the brain. The online identification capability is desirable especially for implantable medical devices \cite{o2018nurip}, for which the learned graphs can serve as a biomarker.
 
\subsection{Methodology}
 
Existing epilepsy works which adopt the network perspective of the brain, use a variety of metrics to show the ability of the inferred networks to distinguish between seizure and non-seizure brain states \cite{shen2019nonlinear,burns2014network,khambhati2017recurring}. For instance, \cite{schindler2006assessing} reports an increase in the eigenvalues of the correlation matrix when in the seizure state. There exist two issues with such state identification in the existing literature.

First, a majority of the existing studies extract a number of scalar metrics from the learned brain graphs (e.g., the eigenvalues). Then, they use each metric, as a single variable in a boxplot, to demonstrate the difference between such metrics in seizure vs. non-seizure states \cite{shen2019nonlinear,burns2014network,khambhati2017recurring}. Since such \textit{univariate} analysis investigates the metrics disjointedly, it is generally less powerful and it reduces the reliability of brain state identification when compared to \textit{multivariate} analysis.

The second issue relates to the real-time implementation of the graph learning algorithms. In some of the existing graph-based literature, computing a single real-time graph from the current signal sample is not possible. Rather, a number of graphs defined from the signal over a longer period of time must be collected to identify the graphs' corresponding brain states \cite{burns2014network,khambhati2015dynamic}. As a result, such methods cannot be implemented in an online manner meaning delays are incurred.

\begin{figure}[!t]
\centering
\includegraphics[width=3.5in]{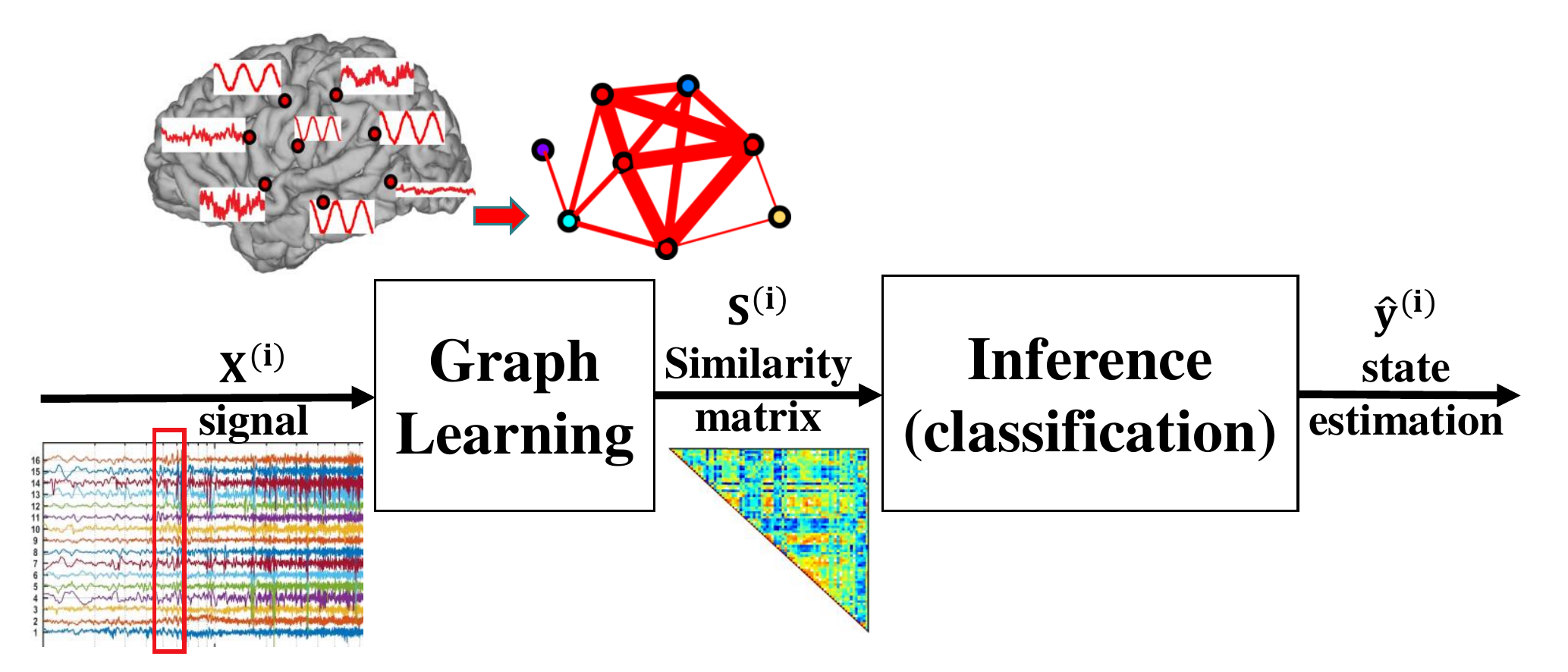}
\caption{Block diagram of the overall system}
\label{fig.system_model}
\end{figure}

In order to address the aforementioned issues, we use machine learning classification (supervised learning) which provides both \textit{multivariate} analysis and an online framework for state identification purposes.

Fig.~\ref{fig.system_model} shows a high-level block diagram of the overall system. In the training phase, training samples are fed to the proposed graph learning method. The embedding ${{Z}}^{(i)}$ and the similarity matrix $S^{(i)}$ corresponding to the signal sample $X^{(i)}$ are computed using the set of parameters $\left\lbrace \Psi^k: k\in[K]\right\rbrace$ and $\mathbf{\Theta}$. Since we use the aggregation functions $g_u^{k,\textrm{mean}}$ and $g_u^{k,\textrm{max}}$ in the experimental results, we set $\Psi^k=\lbrace {U}^{k}, \mathbf{b}^{k}\rbrace$. We choose a \textit{parameter mode}, from three options, for either of the sets of parameters $\Theta$ and $\Psi^k$. The three parameter modes include ``full'', ``diagonal-repeated'' and ``scalar'' which are elaborated on in Appendix~\ref{appendix_parameters}. The introduction of these modes adds flexibility to the number of variables to be optimized, and it leads to dramatic improvements in the results.

To learn the parameters, the optimization problem \eqref{problem_statement} is solved via  mini-batch stochastic gradient descent (SGD). This part of the training phase is done in an unsupervised manner, where sample labels (indicating brain states) are not required. Next, the resulting $S^{(i)}$ are used for classification. Since similarity matrices are symmetric, the elements on the upper (or lower) triangular part of each $S^{(i)}$ are stacked up into a vector. The resulting vectors are then fed to a random forest classifier as the input. The classifier uses the samples' labels for training.  It outputs $\hat{y}^{(i)}$ as the estimate of the $i$th sample's state. Ictal, pre-ictal and inter-ictal respectively refer to the period of seizure occurrence, $10$ seconds before seizure occurrence, and periods of normal brain activity. 

 We label inter-ictal as state $0$ and the other two as state $1$.   Pre-ictal and ictal labels were combined to compensate for potential ambiguities in the precise time of seizure onset \cite{haut2002interrater}. This increases our confidence that our labels include the true time window of seizure onset, which is critical for clinical utility. Although it surely introduces some false “ictal” labels, these should be very small relative the number of true ictal labels and have minimal impact on performance.
 
 \begin{figure}[!t]
\centering
\subfloat[Graph samples using the correlation matrix ]{\includegraphics[width=3.5in]{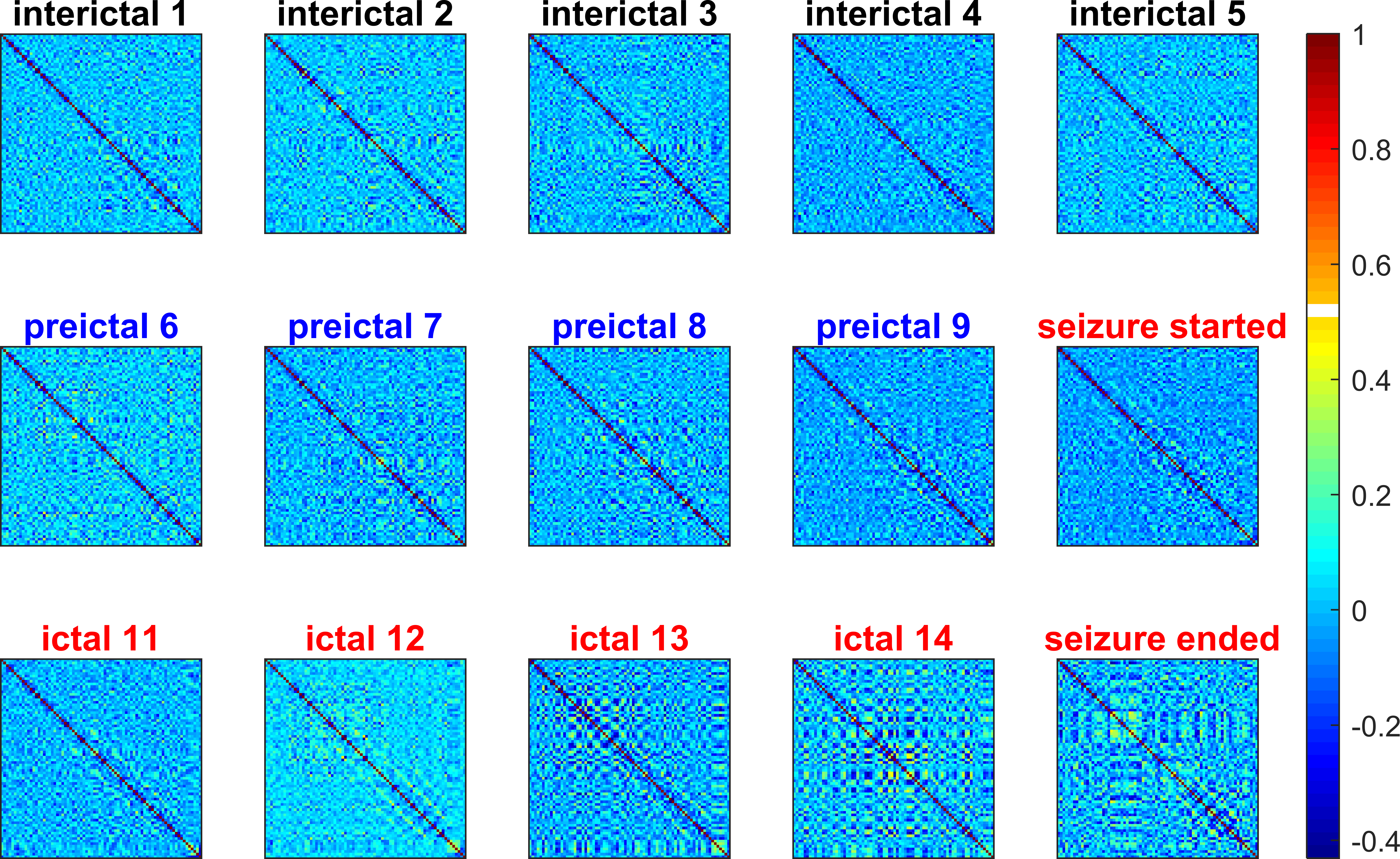}
\label{fig.sample_raw}}
\hfil
\subfloat[Graph samples using NCDD graph learning in the time domain ]{\includegraphics[width=3.5in]{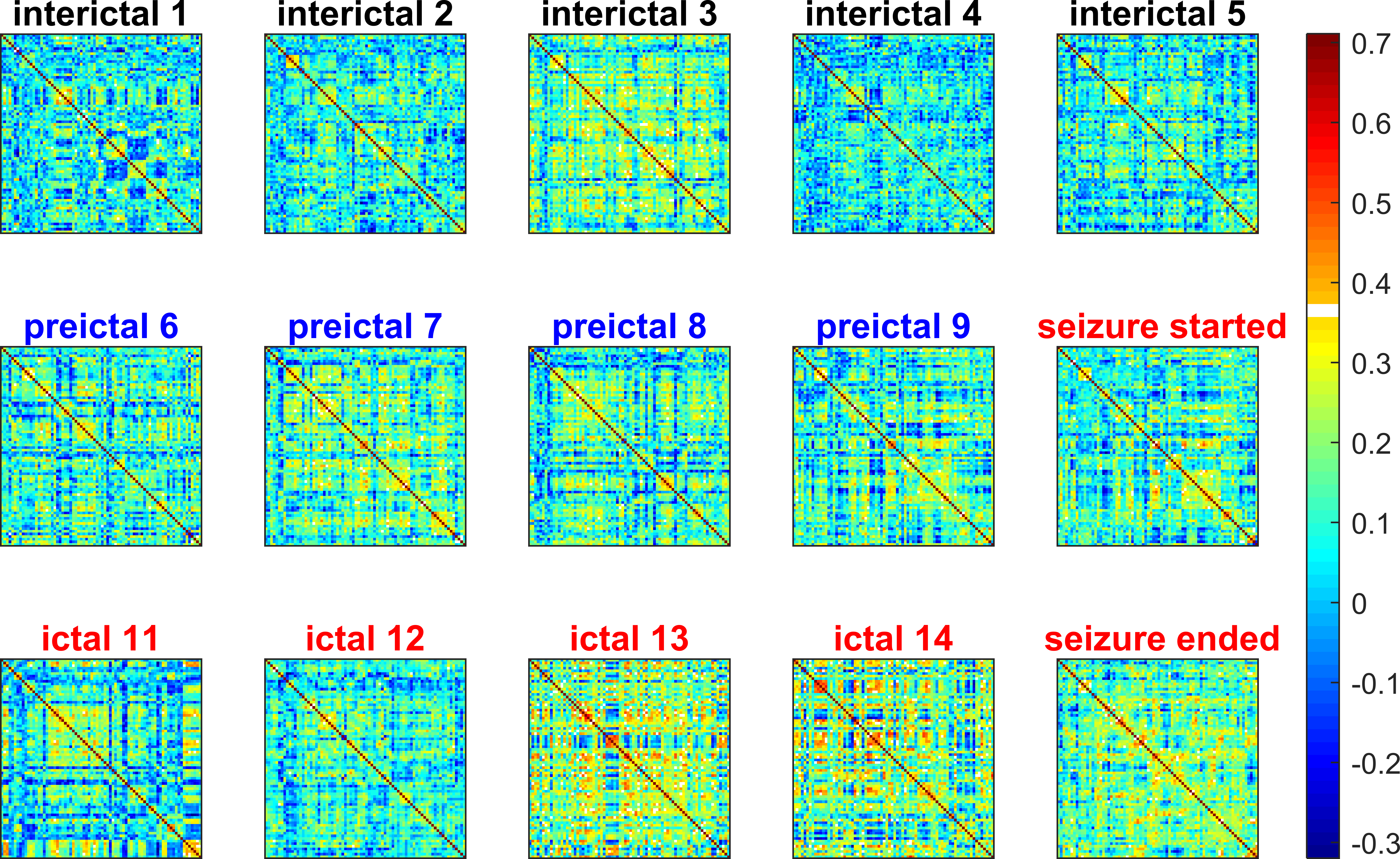}
\label{fig.sample_graphL}}
\caption{ The graphs (the similarity matrices) of normal brain activity in addition to one seizure event in patient $3$ (a hard patient), using two graph learning methods in the time domain. Each sample is the graph computed from a signal window of length $2.5$ seconds. Graphs correspond to the windows of signal sequentially ordered in time, from left to right, and from up to bottom.}
\label{fig.sample_graphs}
\end{figure}
 
\subsection{Data}
 
We used the data from Epilepsiae, the human iEEG epilepsy dataset \cite{EU_dataset}.   The Epilepsiae database contains data from $30$ patients. We used data from a $1/3$ of the patients due to time and computational constraints. We selected the $10$ patients for inclusion in our study because they had good post-surgical outcomes, suggesting that clinician seizure labels were correct, and they had relatively large numbers of seizures, which facilitates training. 

The data for each patient contains ``clips'', which are brain signal recordings over a period of about one-hour. Each clip was temporally subsampled to $256$ Hz and divided into windows of $2.5$-second length with $1.5$-second overlap. Each window is denoted as a signal sample $X^{(i)}\in\mathbb{R}^{N\times T}$. Since the number of inter-ictal samples in the dataset is significantly greater than that of pre-ictal and ictal samples, the data is imbalanced. To address this issue, we used all the samples of state $1$ and randomly subsampled the signals of state $0$, such that the sample ratio of state $0$ to state $1$ is reduced to $10$. We then separately sorted the samples of states $0$ and $1$ chronologically, where the first and second halves of the samples of each state were used for training and testing, respectively.

\begin{table}
\begin{center}
\caption{Constants and hyperparameters in the experiments, shared between patients}
\label{tab:shared_hyperparameters}
\begin{tabular}{|c|c|}
\hline \hline 
\textbf{Hyperparameter/Constant}  &   \textbf{Value}\\
\hline \hline 
 \multicolumn{2}{|c|}{\textbf{Data Processing} }   \\
\hline
length of time before clinical seizure onset & $10$ sec\\
 that is defined as the pre-ictal state \\
\hline
$T$ (length of the signal on each node)  & $640$ \\
\hline
$\tilde{T}$ (number of inner windows) & $3$ \\
\hline
$W$ (number of frequency bins)   & $79$ \\
\hline
\multicolumn{2}{|c|}{\textbf{Graph Learning} }   \\
\hline
$K$ (number of aggregation iterations)  & $1$\\
\hline
$g_u^k$ (aggregation function)  & $g_u^{k,\textrm{mean}}$ \\
\hline
\multicolumn{2}{|c|}{\textbf{Classification}}   \\
\hline
number of trees in the random forest classifier & $1000$ \\
\hline\hline 
\end{tabular}
\end{center}
\end{table}

\begin{table*}
\begin{center}
\caption{Constants and hyperparameters in the experiments, adjusted per patients }
\label{tab:individual_hyperparameters}
\begin{tabular}{|c|c|c|c|c|c|c|c|c|c|c|c|c|c|c|c|}
\hline \hline 
\textbf{Patient} & \textbf{Name in}   & \textbf{Number of}& \textbf{domain}  & \textbf{SGD number } & \textbf{$\eta^{\textrm{ratio}}$} & \textbf{SGD learning} & \textbf{SGD batch} & $\Theta$\textbf{-mode}  & $\Psi^k$\textbf{-mode}\\
\textbf{number}&\textbf{EU dataset}& \textbf{nodes ($N$)}&&\textbf{of epochs}&    & \textbf{rate}  &\textbf{size}& & \\
\hline
$1$  &PT620& $31$& time & $1$ & $0.7$ & $0.1$ & $200$ & scalar & full\\
 && & frequency& $1$ & $0.5$ & $0.1$ & $200$ & scalar & diagonal-repeated\\
\hline
$2$   & PT1125 & $49$& time & $2$ & $0.5$ & $0.1$ & $500$ & scalar & full\\
 && & frequency& $2$ & $0.7$ & $0.1$ & $200$ & scalar & full\\
\hline
$3$  & PT565  & $76$& time  & $1$ & $0.7$ & $0.1$ & $200$ & scalar & full\\
&& & frequency& $3$ & $0.5$ & $0.1$ & $200$ & diagonal-repeated & diagonal-repeated\\
\hline
$4$   & PT958  & $76$ & time& $2$ & $0.7$ & $0.1$ & $200$ & scalar & full\\
& & & frequency& $4$ & $0.5$ & $0.1$ & $200$ & diagonal-repeated & diagonal-repeated\\
\hline
$5$  & PT273 & $35$  & time& $1$ & $0.7$ & $0.1$ & $500$ & scalar & full\\
 && & frequency& $1$ & $0.5$ & $0.1$ & $200$ & diagonal-repeated & diagonal-repeated\\
\hline
$6$  & PT442  & $53$ & time& $2$ & $0.5$ & $0.1$ & $200$ & full & full\\
 && & frequency& $2$ & $0.7$ & $0.001$ & $200$ & diagonal-repeated & diagonal-repeated\\
\hline
$7$  & PT1096 & $61$& time  & $1$ & $0.5$ & $0.1$ & $200$ & full & full\\
 & & &frequency& $1$ & $0.7$ & $0.01$ & $200$ & scalar & diagonal-repeated\\
\hline
$8$  & PT590 & $79$ & time & $1$ & $0.7$ & $0.1$ & $200$ & scalar & scalar\\
 && & frequency& $1$ & $0.6$ & $0.1$ & $200$ & diagonal-repeated & diagonal-repeated\\
\hline
$9$  & PT970  & $89$& time & $1$ & $0.5$ & $0.1$ & $200$ & full & full\\
 && & frequency& $2$ & $0.5$ & $0.1$ & $200$ & diagonal-repeated & diagonal-repeated\\
\hline
$10$  & PT1077  & $99$& time  & $1$ & $0.5$ & $0.1$ & $200$ & full & full\\
 && & frequency& $1$ & $0.7$ & $0.1$ & $200$ & diagonal-repeated & diagonal-repeated\\
\hline \hline
\end{tabular}
\end{center}
\end{table*}

\begin{figure}[!t]
\centering
\includegraphics[width=3.5in]{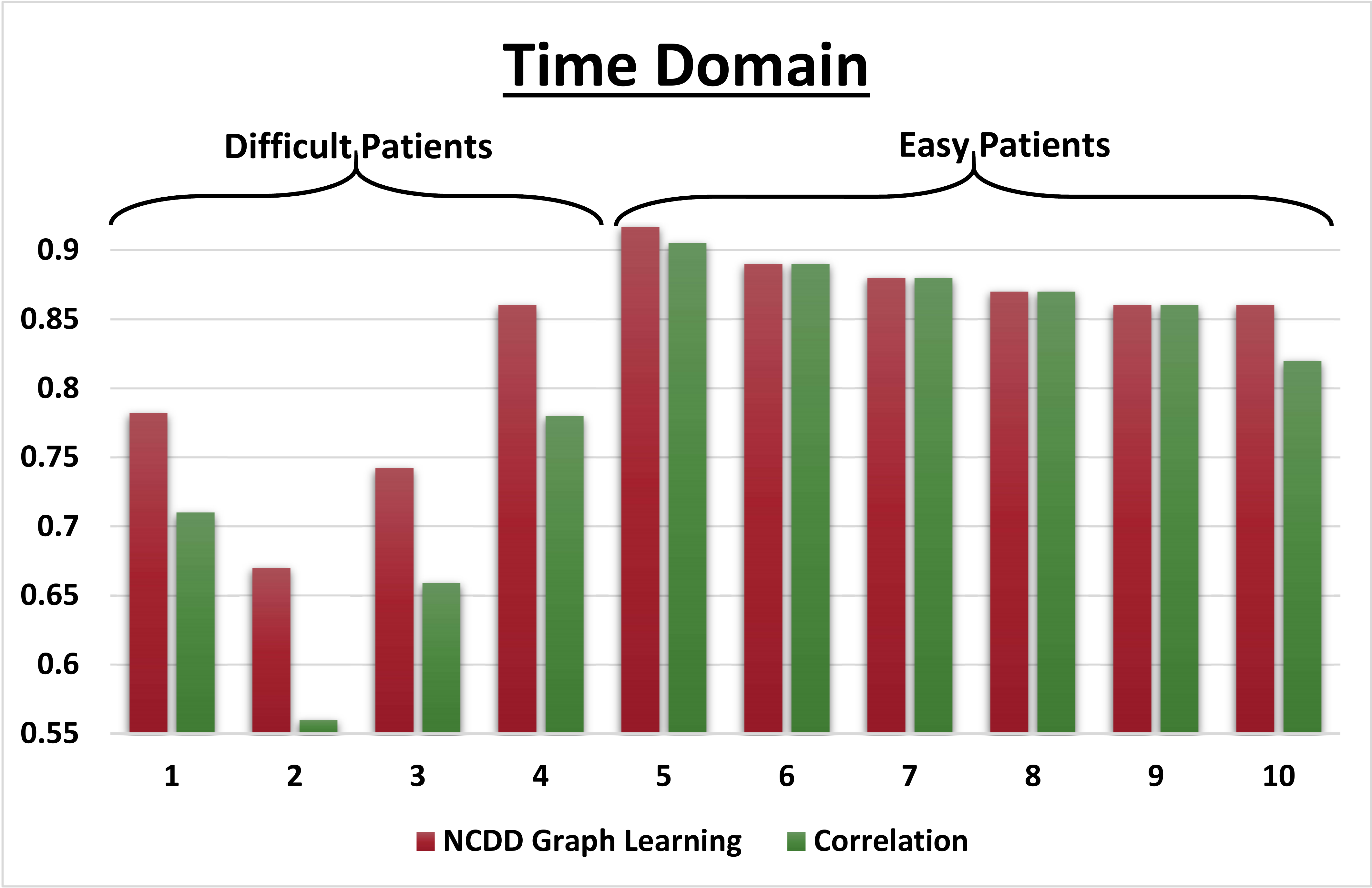}
\caption{ AUC measure, comparing the seizure detection classification performance of the graph learning methods in the time domain, on the testing signal samples of $10$ patients. Note that the vertical axis starts at $0.55$.}
\label{fig.time_domain_results}
\end{figure}

\begin{remark}
The alternative to the temporally disjoint ``train-test sample selection scheme''  is to use ``random'' sets that disregard time. For example in the former case we may train on the first $5$ days and test on the next $5$ days. However, in the latter case we may, for instance, train on days $2,4,6,7,8$ and then test on days $1,3,5,9,10$. We pick to use the temporally disjoint approach for two reasons. First, the disjoint approach mimics the real life situation where data is collected contiguously, used for training, and then applied to future data.  Second, the temporally disjoint scenario means that training and testing samples are, on average, temporally the furthest from each other. There may exist a gradual change in the features of brain activity (medication changes, ongoing seizures, etc.). Hence, the temporally disjoint scenario gives a lower bound (worst case) on the performance, compared to other train-test sample selection methods.
\end{remark}
 
\subsection{Results}

Fig. \ref{fig.sample_graphs} illustrates the similarity matrices computed using both the correlation matrix and our NCDD graph learning algorithm in the time domain.

The whole system illustrated in Fig.~\ref{fig.system_model} was implemented in Python using Tensorflow. We built on and made various changes to the GraphSAGE implementation in Python \cite{graphsage_code}. Table \ref{tab:shared_hyperparameters} lists the main functions and the hyperparameters used in the experiments that were shared between the patients. The main hyperparameters that were adjusted per patient, are listed in Table~\ref{tab:individual_hyperparameters}. Since the scale of the values of $\frac{1}{I}\displaystyle\sum_{i\in\mathcal{I}} \matidx{(P^{(i)})^{-1}}{u,v}$  in  \eqref{graph_topology} may differ between patients, it is easier to determine $\eta$ such that a level of sparsity for $A$ is achieved. We therefore used the parameter $\eta^{\textrm{ratio}}$ to determine the fraction of zero elements in $A$. For instance, if  $\eta^{\textrm{ratio}}=0.5$, the parameter $\eta$ in \eqref{graph_topology} is determined from the matrices $(P^{(i)})^{-1}$ such that half of the $A$ elements would be zero.

In order to have a fair comparison, we used the correlation matrix as a baseline to compare our graph learning method in the time domain. Similarly, the non-normalized coherence matrix was the baseline to evaluate the results of the NCDD graph learning in the frequency domain. The evaluation procedure was as follows. The samples were passed through four graph learning methods: correlation, non-normalized coherence, NCDD in the time domain, and NCDD in the frequency domain. For each method, the outcome consisted of matrices used as the classifier input. The samples used to train the graph learning were also used for training the classifier. The rest of the samples were tested both in the  classification and graph learning modules, with and without their corresponding labels. The area under the receiver operating characteristic curve (AUC) is a scalar evaluation metric in binary classification; it equals $1$ and $0.5$ for a perfect and random classification, respectively.

 We divided the patients into difficult and easy categories. 
 
\begin{figure}[!t]
\centering
\includegraphics[width=3.5in]{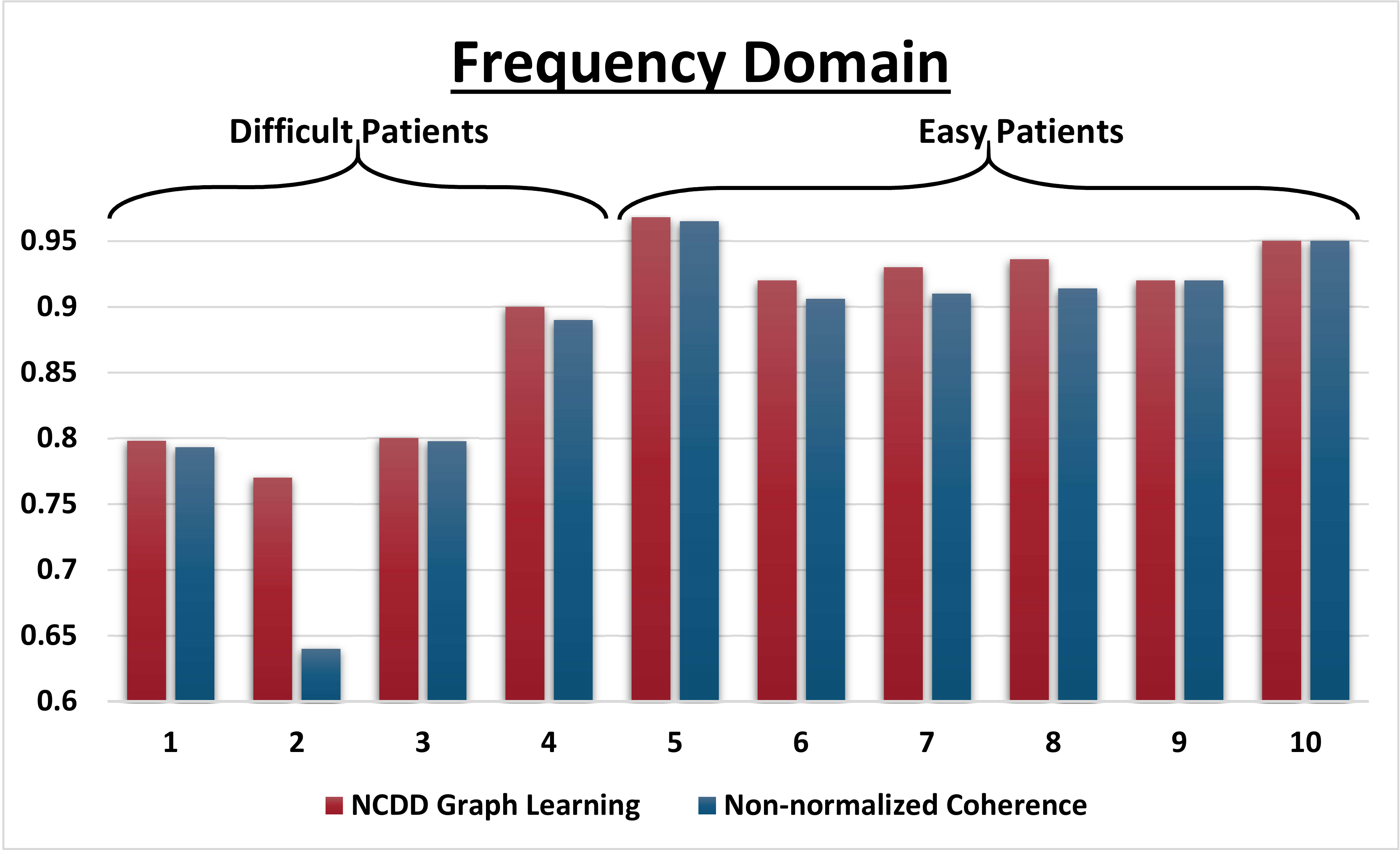}
\caption{AUC measure, comparing the seizure detection classification performance of the graph learning methods in the frequency domain, on the testing signal samples of $10$ patients.}
\label{fig.freq_domain_results}
\end{figure}

The intuition behind such division was as follows. Looking at figures \ref{fig.time_domain_results} and \ref{fig.freq_domain_results}, we observed that patients $1$ to $4$ had the lowest AUC, \textit{both} in the time and frequency domains. We termed these four patients the ``difficult'' patients, and the rest of them, the ``easy'' patients. In other words, for the difficult patients, it was hard to distinguish seizure from non-seizure samples.

Fig. \ref{fig.time_domain_results} shows the classification results in the time domain, where it is observed that our proposed  graph learning method in the time domain improved  binary classification results by $8.63$ percent for the difficult patients. 
Fig. \ref{fig.freq_domain_results} shows the classification results in the frequency domain, in which the average improvement of our proposed  graph learning method is $3.68$ percent for the difficult patients. 
As observed from both figures \ref{fig.time_domain_results} and \ref{fig.freq_domain_results}, the improvement in AUC of easy patients, in both domains, is small.

In order to combine the results in the two domains, we took  the maximum of the AUC measure in the two domains, per patient. The combined improvement was $9.13$ percent for the difficult patients. 

\begin{figure}[!t]
\centering
\includegraphics[width=3.5in]{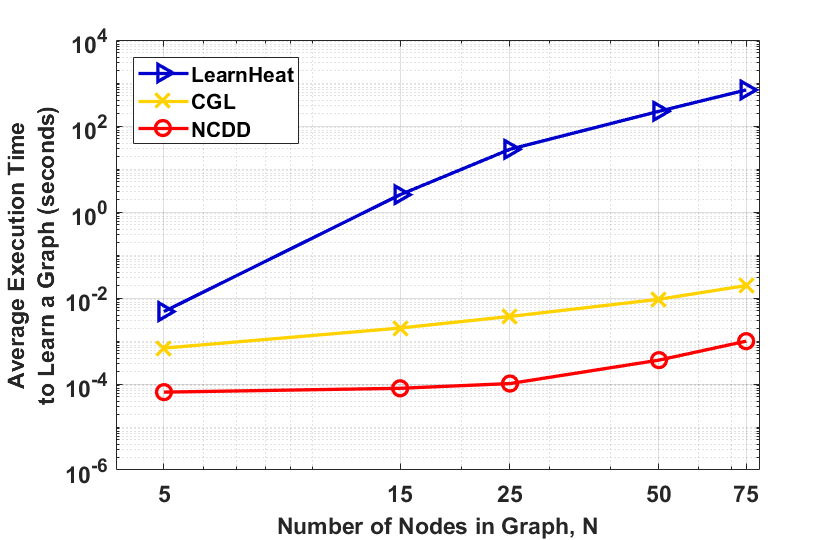}
\caption{Comparing the complexity (average execution time) of NCDD,  LearnHeat \cite{thanou2017learning}, and CGL \cite{egilmez2017graph}. Note that both axes are in log scale.}
\label{fig:comp_complexity}
\end{figure}

As mentioned in Sec.~\ref{sec:preliminaries}, one of the main features of NCDD graph learning is its scalability. Equations \eqref{EC_per}, \eqref{XtoZ}, \eqref{ZtoS} and the discussions following suggest that the complexity of graph learning using NCDD is less than the complexity of edge-centric methods. To confirm this conjecture, we evaluated the execution time per graph computation for NCDD, the LearnHeat algorithm of \cite{thanou2017learning}, and the CGL algorithm of \cite{egilmez2017graph}. We generated $100$ i.i.d. samples of graph signal $X$ per equation \eqref{x_def}. To have a fair comparison and to avoid bias in the results towards a graph that might perform better with a particular graph learning algorithm, each $X$ is generated without assuming a particular underlying graph. In particular, we generate each element of each sample matrix $X$, randomly and independently with a uniform distribution between $0$ and $1$. We fix the feature size $T=50$ and sweep over variable $N \in {5, 15, 25, 50, 75}$. For each graph learning method, we averaged over the execution time to learn the graph from each of $100$ samples of $X$, for each value of $N$. The results are plotted in Fig.~\ref{fig:comp_complexity}. The x and y-axes correspond to the number of nodes and the execution time in seconds, respectively. Figure \ref{fig:comp_complexity} demonstrates that NCDD speeds up the graph learning process up to $\mathcal{O}(10^6)$ times when compared to LearnHeat, and it performs $\mathcal{O}(10)$ times faster than CGL. 

%
 
\section*{Acknowledgments}
The authors would like to thank Dr. Thanou et al. and Egilmez et al. for sharing the MATLAB implementation of their graph learning algorithms in \cite{thanou2017learning} and \cite{egilmez2017graph}, respectively.

\section{Conclusions} \label{sec:conclusion}
We have developed a scalable node-centric graph learning method based on  representation learning on graphs. We applied the method to model time-varying brain networks. The scheme's superiority in differentiating brain state when compared to conventional schemes in network neuroscience is confirmed for seizure detection purposes on a dataset of $10$ patients. In future work, one could explore the connection between the proposed node-centric approach and existing edge-centric methods. Furthermore, one may also wish to investigate the expressive power of the proposed method to model graph signals. We intend to extend the developed technique to other problems of brain state identification like working memory behaviour.

\appendices
\section{Modes of Parameter Definition}\label{appendix_parameters}
In Sec.~\ref{sec:results}, we used three modes for each of the sets of parameters $\mathbf{\Theta}$ and $\lbrace {U}^{k}, \mathbf{b}^{k}\rbrace$. The modes refer to different scenarios where different number of variables are optimized over in the optimization problem of \eqref{problem_statement}.
\begin{enumerate}
    \item \textit{Full}: In this mode, all the elements in the original parameter sets, are optimization variables. They include  $U^k\in\mathbb{R}^{D_0\times D_0}$; $\mathbf{b}^k\in\mathbb{R}^{D_0}$; and $\mathbf{\theta}\in\mathbb{R}^{D}$ in the time domain and $\left\lbrace\mathbf{\theta}^\twoparts\in\mathbb{R}^{W}:\alpha\in\lbrace a,b\rbrace\right\rbrace$ in the frequency domain.
    \item \textit{Diagonal-repeated}: In this mode, which is only defined in the frequency domain, we first divide the set of frequency bin values $\Upsilon$ to six physiologically signified signal bands namely: $\delta$ (0.1-4Hz), $\theta$ (4-8Hz), $\twoparts$ (8-13Hz), $\beta$ (13-30Hz), $\gamma$ (30-50Hz), and high-$\gamma$ (70-100Hz). We then give the same weight to the frequency bins in the same band. To accomplish this, we first define base variable vectors $\tilde{\mathbf{\theta}}^\twoparts, \tilde{\mathbf{u}}, \tilde{\mathbf{b}} \in  \mathbb{R}^6$. We assume $j_1,j_2,\cdots,j_6$ denote the number of frequency bins in the six bands, where $\sum_{l=1}^6 j_l=W$. We then define the parameter sets by repeating these variables in the following form:
    \begin{align}
        \nonumber
        &\mathbf{\theta}^\twoparts =
            \left[\begin{array}{c}
                \tilde{\theta}^\twoparts_1  \mathbf{1}_{j_1} \\\tilde{\theta}^\twoparts_2 \mathbf{1}_{j_2}\\ \vdots  \\ \tilde{\theta}^\twoparts_6 \mathbf{1}_{j_6} \\
            \end{array}\right],
            \mathbf{b}^k = 
           \left[ \begin{array}{c}
                \tilde{b}_1 \mathbf{1}_{j_1} \\\tilde{b}_2 \mathbf{1}_{j_2}\\ \vdots \\ \tilde{b}_6 \mathbf{1}_{j_6}\\
            \end{array}\right],\\
            &U^k=f^{\textrm{diag}}\left(\left[
            \begin{array}{c}
                \tilde{u}_1 \mathbf{1}_{j_1} \\\tilde{u}_2 \mathbf{1}_{j_2} \\ \vdots \\ \tilde{u}_6 \mathbf{1}_{j_6} \\
            \end{array}
            \right]\right).
    \end{align}
    \item \textit{Scalar}: Only one scalar variable per array (or matrix) is used in this mode, i.e., using $\dbtilde{b},\dbtilde{u},\dbtilde{\theta},\dbtilde{\theta}^\twoparts\in\mathbb{R}$ we define:
    \begin{align}
        \nonumber
        &\mathbf{b}^k = \dbtilde{b}\mathbf{1}_{D_0}\\
        &U^k=\dbtilde{u}\mathbf{1}_{D_0\times D_0}.
    \end{align}
    Also, in the time and frequency domains we have:
    \begin{align}
        \nonumber
        &\mathbf{\theta} = \dbtilde{\theta}\mathbf{1}_{D}\\
        &\mathbf{\theta}^\twoparts = \dbtilde{\theta}^\twoparts\mathbf{1}_{W},
    \end{align}
    where the matrix ${1}_{D_0\times D_0}$ is an all-one matrix of size $D_0\times D_0$.
\end{enumerate}

\ifCLASSOPTIONcaptionsoff
  \newpage
\fi

\bibliographystyle{IEEEtran}
\bibliography{IEEEabrv,references}
\newpage
\begin{IEEEbiography}
    [{\includegraphics[width=1in,height=1.25in,clip,keepaspectratio]{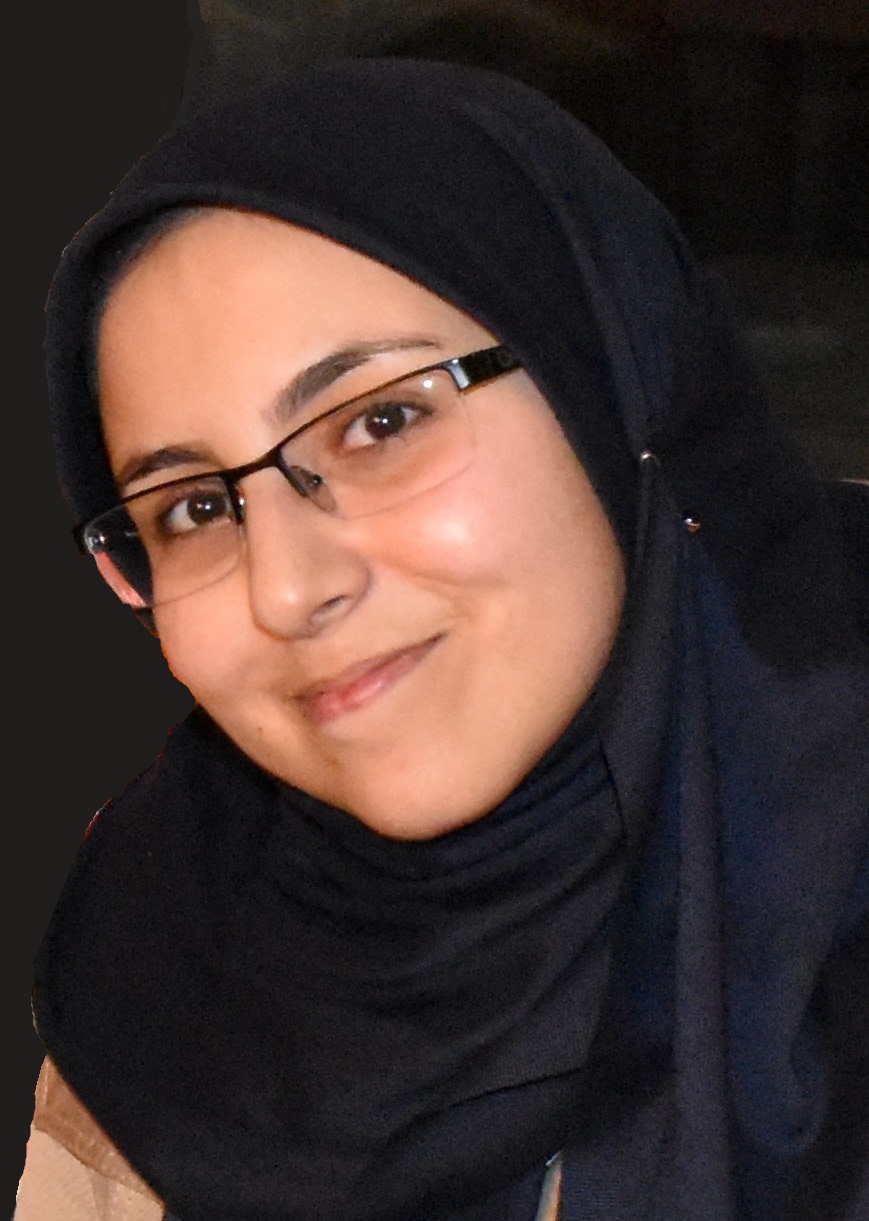}}]{Nafiseh Ghoroghchian}
 received the B.Sc. and M.Sc. degrees in electrical engineering from Sharif University of Technology, Iran, in 2015 and 2017, respectively. She is currently studying PhD. in Electrical and Computer Engineering at the University of Toronto, Canada. Her research interests include signal processing and developing learning algorithms. She ranked 10 (among 350,000 participants) in the Iranian University Entrance Exam. She was a member of Iran’s National Elite Foundation and received a six-year fellowship. Since 2017, she has been the recipient of Connaught International Scholarship for Doctoral students.
\end{IEEEbiography}

\begin{IEEEbiography}
    [{\includegraphics[width=1in,height=1.25in,clip,keepaspectratio]{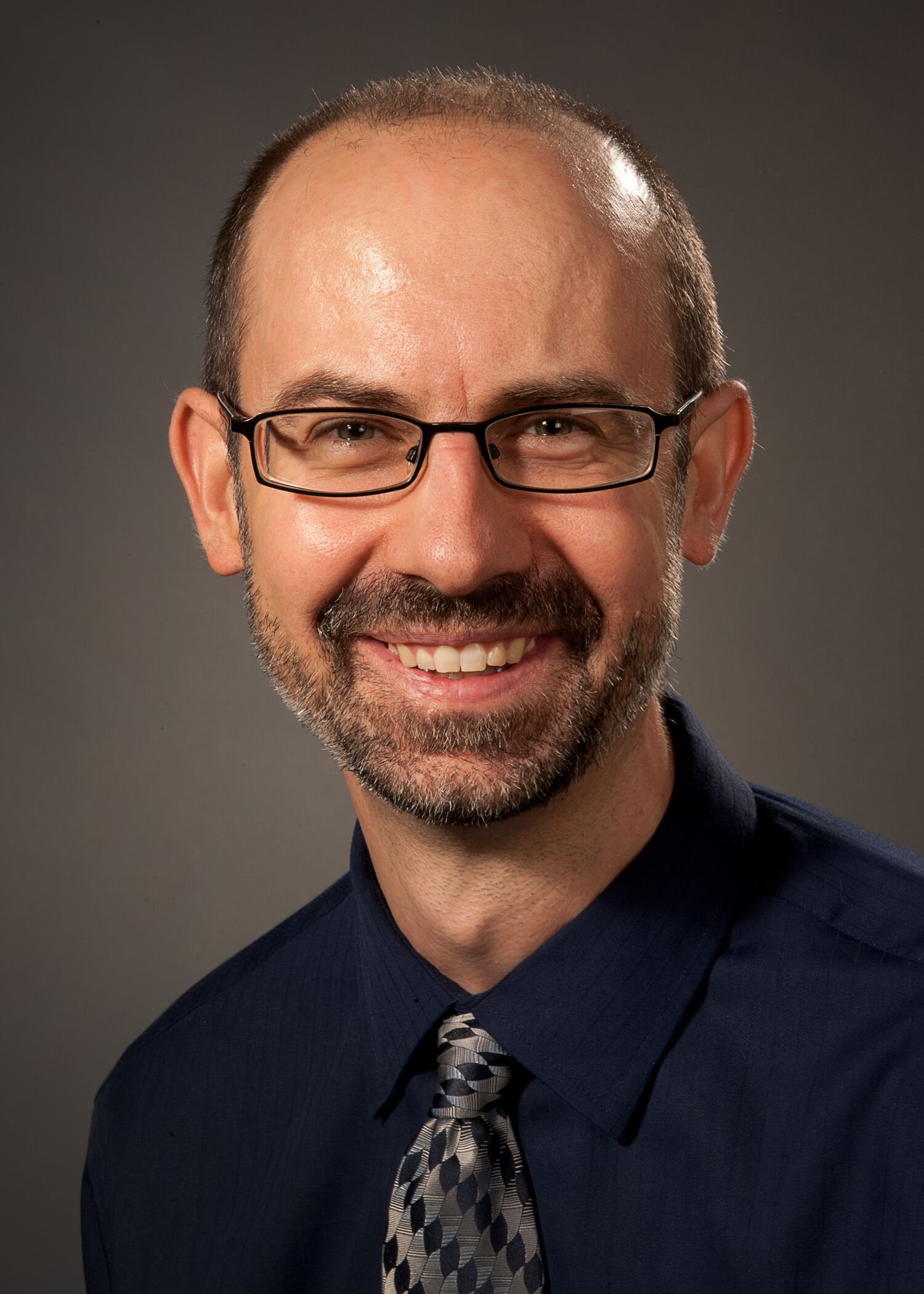}}]{David Groppe}
    is a data scientist who specializes in the analysis of electroencephalogram (EEG) data and epilepsy at the Krembil Research Institute in Toronto. He has published widely on statistical methods for EEG analysis and is lead developer of two open-source EEG analysis software packages, the Mass Univariate ERP Toolbox and iELVis. Dr. Groppe’s reviewer of mass univariate analyses of event-related potentials (ERPs) is listed as one of the 10 papers every new ERP researcher should read, by the prominent ERP scientist Steven Luck. Dr. Groppe also published the first quantitative atlas of intracranial EEG brain rhythms in 2013.
\end{IEEEbiography}

\begin{IEEEbiography}
    [{\includegraphics[width=1in,height=1.25in,clip,keepaspectratio]{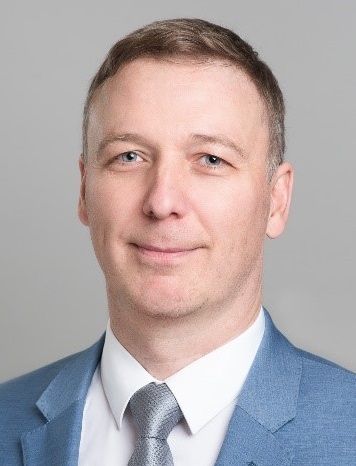}}]{Roman Genov} (S’96–M’02–SM’11) received the B.S. degree in Electrical Engineering from Rochester Institute of Technology, NY in 1996 and the M.S.E. and Ph.D. degrees in Electrical and Computer Engineering from Johns Hopkins University, Baltimore, MD in 1998 and 2003 respectively.
    
    He is currently a Professor in the Department of Electrical and Computer Engineering at the University of Toronto, Canada, where he is a member of Electronics Group and Biomedical Engineering Group and the Director of Intelligent Sensory Microsystems Laboratory. Dr. Genov’s research interests are primarily in analog integrated circuits and systems for energy-constrained biological, medical, and consumer sensory applications.
     
    Dr. Genov is a co-recipient of Jack Kilby Award for Outstanding Student Paper at IEEE International Solid-State Circuits Conference, Best Paper Award of IEEE TRANSACTIONS ON BIOMEDICAL CIRCUITS AND SYSTEMS, Best Paper Award of IEEE Biomedical Circuits and Systems Conference, Best Student Paper Award of IEEE International Symposium on Circuits and Systems, Best Paper Award of IEEE Circuits and Systems Society Sensory Systems Technical Committee, Brian L. Barge Award for Excellence in Microsystems Integration, MEMSCAP Microsystems Design Award, DALSA Corporation Award for Excellence in Microsystems Innovation, and Canadian Institutes of Health Research Next Generation Award. He was a Technical Program Co-chair at IEEE Biomedical Circuits and Systems Conference, a member of IEEE European Solid-State Circuits Conference Technical Program Committee, and a member of IEEE International Solid-State Circuits Conference International Program Committee. He was also an Associate Editor of IEEE TRANSACTIONS ON CIRCUITS AND SYSTEMS-II: EXPRESS BRIEFS and IEEE SIGNAL PROCESSING LETTERS, as well as a Guest Editor for IEEE JOURNAL OF SOLID-STATE CIRCUITS. Currently he is an Associate Editor of IEEE TRANSACTIONS ON BIOMEDICAL CIRCUITS AND SYSTEMS.
\end{IEEEbiography}

\vfill\vfill\vfill\vfill\vfill\vfill
\begin{IEEEbiography}
    [{\includegraphics[width=1in,height=1.25in,clip,keepaspectratio]{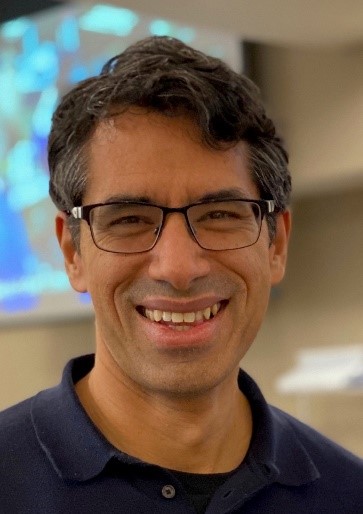}}]{Dr. Taufik A Valiante} MD PhD FRCS is an Associate Professor of Neurosurgery at the University of Toronto, with cross-appointments to Electrical and Computer Engineering, and the Institute of Biomaterials and Biomedical Engineering.  His surgical specialization is in Epilepsy Surgery, and he directs the Surgical Epilepsy Program at the Krembil Neuroscience Center at the Toronto Western Hospital where he holds his clinical appointment.  He is a scientist at the Krembil Research Institute, using electrophysiology, behavior, and genomics to understand the physiology and pathology of the human brain. Additionally he Co-Directs CRANIA (Center for Advancing Neurotechnological Innovation to Application) with the view of translating basic neuroscience towards therapeutic benefit using neuromodulation.
\end{IEEEbiography}

\begin{IEEEbiography}[{\includegraphics[width=1in,height=1.25in,clip,keepaspectratio]{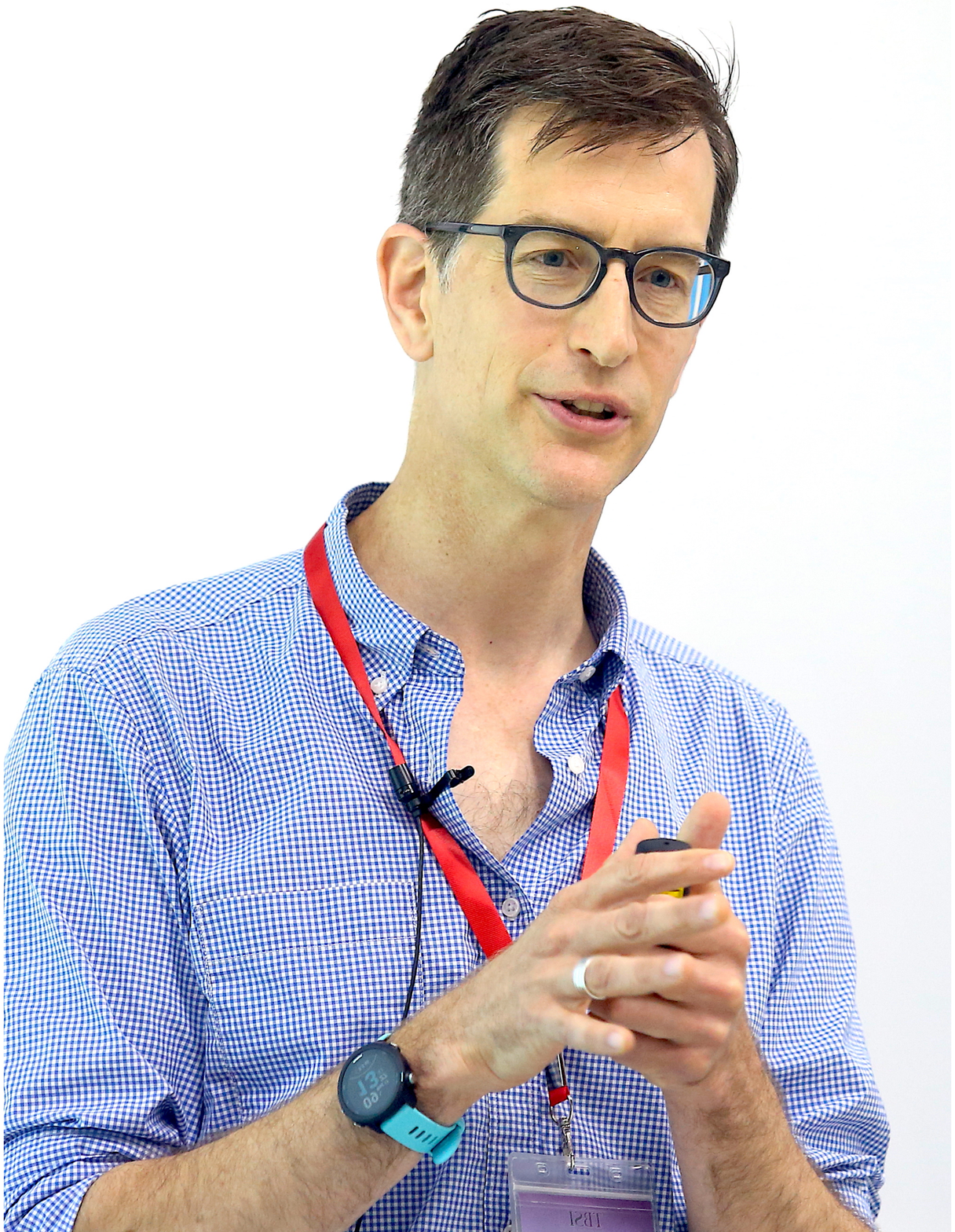}}]{Stark C. Draper} (S'99-M'03-SM'15) is a  Professor of Electrical and Computer Engineering at the University of Toronto.  He received his undergraduate degrees (BS in EE and BA in history) from Stanford University and his MS and PhD degrees in EECS from MIT.  Dr. Draper completed postdocs at the University of Toronto and at the University of California, Berkeley.  
He then worked at the Mitsubishi Electric Research Labs (MERL).  Before returning to Toronto he was 
an assistant and associate professor at the University of Wisconsin, Madison.  Professor Draper’s 
research interests include information and coding theory, optimization and security, and the application 
of these disciplines to problems in communications, computing, and learning.  Recent industrial 
collaborative and consulting positions include with Huawei, AMD, Disney Research, and MERL.  
Dr. Draper chairs the new “Machine Intelligence” major at UofT 
and serves on the IEEE Information Theory Society Board of Governors.  He is spending the 
2019-20 academic year on sabbatical visiting the Chinese University of Hong Kong, Shenzhen.
\end{IEEEbiography}
\vfill\vfill\vfill\vfill\vfill\vfill
\end{document}